\documentclass[final]{cvpr}

\usepackage{times}
\usepackage{epsfig}
\usepackage{graphicx}
\usepackage{amsmath}
\usepackage{amssymb}
\usepackage{amsfonts}
\usepackage{url}
\usepackage{algorithm}
\usepackage{algorithmicx}
\usepackage{algpseudocode}
\usepackage{booktabs}
\usepackage{threeparttable}
\usepackage{multirow}
\usepackage{subcaption}
\usepackage{caption}
\usepackage{array}
\usepackage{color}
\usepackage{soul}
\usepackage{footnote}
\usepackage{tablefootnote}

\graphicspath{{figure/}}

\usepackage[pagebackref=true,breaklinks=true,colorlinks,bookmarks=false]{hyperref}

\begin{document}

\title{DS-UI: Dual-Supervised Mixture of Gaussian Mixture Models\\ for Uncertainty Inference}

\author{Jiyang~Xie$^1$, Zhanyu~Ma$^1$, Jing-Hao~Xue$^2$, Guoqiang~Zhang$^3$, and~Jun~Guo$^1$\\
$^1$ Pattern Recognition and Intelligent Systems Lab., Beijing University of Posts and Telecommunications, China.\\
$^2$ Department of Statistical Science, University College London, United Kingdom.\\
$^3$ School of Electrical and Data Engineering, University of Technology Sydney, Australia.\\
{\tt\small $\{$xiejiyang$2013$, mazhanyu, guojun$\}$@bupt.edu.cn, jinghao.xue@ucl.ac.uk, guoqiang.zhang@uts.edu.au}
}

\maketitle

\begin{abstract}
  This paper proposes a dual-supervised uncertainty inference (DS-UI) framework for improving Bayesian estimation-based uncertainty inference (UI) in deep neural network (DNN)-based image recognition. In the DS-UI, we combine the classifier of a DNN, i.e., the last fully-connected (FC) layer, with a mixture of Gaussian mixture models (MoGMM) to obtain an MoGMM-FC layer. Unlike existing UI methods for DNNs, which only calculate the means or modes of the DNN outputs' distributions, the proposed MoGMM-FC layer acts as a probabilistic interpreter for the features that are inputs of the classifier to directly calculate the probability density of them for the DS-UI. In addition, we propose a dual-supervised stochastic gradient-based variational Bayes (DS-SGVB) algorithm for the MoGMM-FC layer optimization. Unlike conventional SGVB and optimization algorithms in other UI methods, the DS-SGVB not only models the samples in the specific class for each Gaussian mixture model (GMM) in the MoGMM, but also considers the negative samples from other classes for the GMM to reduce the intra-class distances and enlarge the inter-class margins simultaneously for enhancing the learning ability of the MoGMM-FC layer in the DS-UI. Experimental results show the DS-UI outperforms the state-of-the-art UI methods in misclassification detection. We further evaluate the DS-UI in open-set out-of-domain/-distribution detection and find statistically significant improvements. Visualizations of the feature spaces demonstrate the superiority of the DS-UI.\footnote{Under review.}
\end{abstract}


\begin{figure}[!t]
    \begin{center}
        \includegraphics[width=0.99\linewidth]{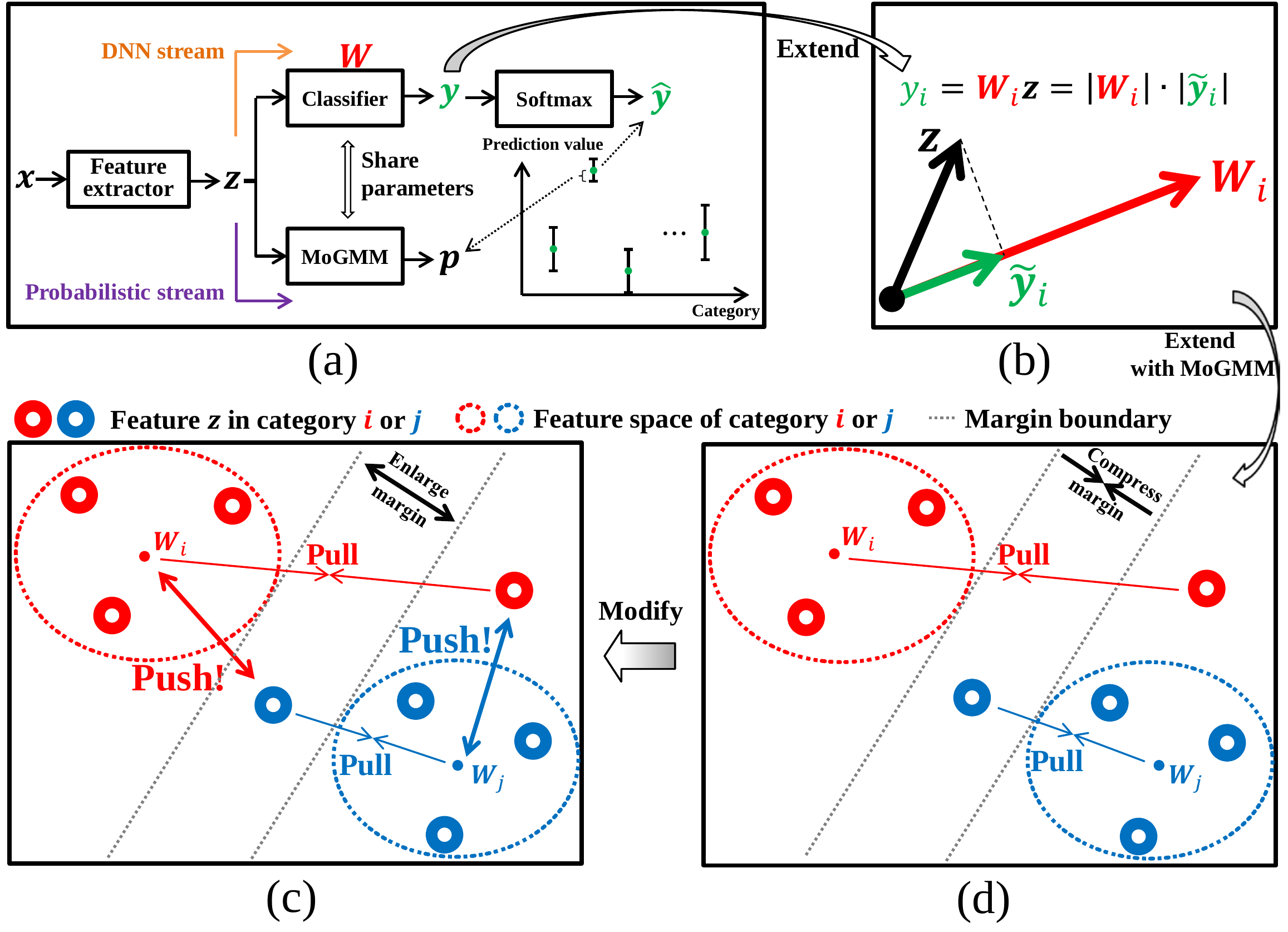}
    \end{center}
    \vspace{-8mm}
    \caption{Illustration of the DS-UI. A conventional DNN architecture for image recognition with cross-entropy loss (orange stream in (a)) can be divided into a feature extractor and a classifier (\emph{i.e.}, the last FC layer) with parameter matrix $\boldsymbol{W}$. Here, $\boldsymbol{W}_i$, the $i^{th}$ row vector of $\boldsymbol{W}$, is assumed to be the center of the $i^{th}$ class. Given a feature vector $\boldsymbol{z}$, the output value $y_i$ of the classifier is proportional to the projection length $|\tilde{\boldsymbol{y}}_i|$ of $\boldsymbol{z}$ in the direction of $\boldsymbol{W}_i$ in (b). We extend the DNN to a model with a parameter-shared MoGMM by adding a probabilistic stream to model $\boldsymbol{z}$ \emph{w.r.t.} the class centers (purple stream in (a)). Traditional SGVB for mixture models and optimization algorithms in other UI methods aim at decreasing the distances between each sample and its corresponding class center, which may undesirably compress the margin between two classes in the case in (d). We modify it by ``pulling'' the positive samples and ``pushing'' the  negative samples simultaneously for the class centers (in (c)) to reduce intra-class distances and enlarge inter-class margins simultaneously by proposing a dual-supervised SGVB, which benefits the DS-UI.}
    \label{fig:motivation}
    \vspace{-4mm}
\end{figure}

\section{Introduction}\label{sec:intro}

Deep neural networks (DNNs) usually tend to output certain and even overconfident predictions for decision, rather than confidence intervals of the predictions~\cite{lakshminarayanan2017simple,malinin2018predictive}. In this case, the DNNs are not able to assess the uncertainty of their outputs. In some tasks, such as medical image analysis and autonomous driving, outputs without uncertainty indications may be catastrophic and limit the applications.

To address the above issue, uncertainty inference (UI) has been introduced for estimating how uncertain the outputs of a DNN are to further improve its reliability and applicability. Generally speaking, uncertainty can be divided into model uncertainty and data uncertainty~\cite{malinin2018predictive}. The former one focuses on measuring the uncertainty in model generalization, such as misclassification detection~\cite{malinin2018predictive}. The latter one aims to measure the uncertainty of the data, including out-of-domain and out-of-distribution detection for noisy data~\cite{kong2020sde}, which are more challenging tasks. The misclassification and the out-of-domain/-distribution detection tasks are the most important tasks in the UI.

Recent works~\cite{gal2016dropout,hendrycks2017a,maddox2019a,malinin2018predictive,malinin2019reverse} on the UI intended to approximate the outputs of DNNs by distributions, such as Gaussian~\cite{gal2016dropout,maddox2019a}, Dirichlet~\cite{malinin2018predictive,malinin2019reverse}, and softmax~\cite{hendrycks2017a} distributions and define uncertainty only on the outputs. In practice, they only calculate the means or modes of the distributions, although in different ways, for the UI.

In this paper, a dual-supervised uncertainty inference (DS-UI) framework is introduced for improving Bayesian estimation-based UI. In the DS-UI, we propose to combine a mixture of Gaussian mixture models (MoGMM) with a fully-connected (FC) layer as an MoGMM-FC layer to replace the classifier of a DNN and calculate probability density of the outputs for the DS-UI directly. In general, a DNN architecture for image recognition can be divided into two cascaded parts,~\emph{i.e.}, a feature extractor that contains multiple convolutional and/or FC layers, and a classifier (the last FC layer), as shown in Figure~\ref{fig:motivation}(a)~\cite{kong2020sde,perera2020generative}. As the Gaussian distribution is a simple and generic distribution, and a mixture of mixture models~\cite{malsiner2017identifying,zio2007a} can better estimate large intra-class variability in complex scenes, we adopt an MoGMM to model both the intra-class variability and the inter-class difference, and accordingly extend the DNN model to a new model with the proposed MoGMM-FC layer for modeling the features \emph{w.r.t.} the class centers (\emph{i.e.}, the row vectors of the parameter matrix of the classifier) and enhancing the learning ability of the classifier. In the MoGMM-FC layer, each Gaussian mixture model (GMM) is learned for one class~\cite{ma2011bayesian,ma2020insights}. Each class center is shared with the weighted summation of the means of the components in the associated GMM and optimized with the MoGMM.

Moreover, traditional stochastic gradient-based variational Bayes (SGVB) algorithms generally supervise the optimization of mixture models by using only positive samples of each class and aim at reducing the distances between samples and their corresponding class centers~\cite{ma2011bayesian,ma2020insights}. However, the margin between different classes might be compressed (see Figure~\ref{fig:motivation}(d)), which may have a negative impact on the performance and can be also found in the optimizations of other UI methods. In this paper, we propose to improve the SGVB for the DS-UI by comprehensively considering both the positive samples (in the class) and the negative samples (in other classes) for each GMM, a strategy defined hereafter as dual-supervised optimization, to reduce the intra-class distances and enlarge the inter-class margins simultaneously, as shown in Figure~\ref{fig:motivation}(c).

The contributions of this paper are four-fold:
\begin{itemize}
    \item A DS-UI framework is introduced. We propose an MoGMM-FC layer with a parameter-shared and jointly-optimized MoGMM to act as a probabilistic interpreter for the features of DNNs to calculate probability density for the DS-UI directly.

    \item We propose a dual-supervised SGVB (DS-SGVB) for the MoGMM-FC layer optimization in the DNNs. The DS-SGVB can enhance the learning ability of the MoGMM-FC layer for the DS-UI.

    \item The proposed DS-UI outperforms the state-of-the-art UI methods in the misclassification detection.

    \item We extend the evaluation of the DS-UI to open-set out-of-domain/-distribution detection (detecting unknown samples from unknown classes or noisy samples) and find statistically significant improvements.
\end{itemize}

\section{Related Work}\label{sec:relatedwork}

\subsection{Uncertainty Inference}\label{ssec:ui}

Most of the recently proposed UI methods define uncertainty only on the outputs and calculate the means or modes of the DNN outputs' distributions.

Blundell~\emph{et al.}~\cite{blundell2015weight} proposed a backpropagation-compatible algorithm with unbiased MC gradients for estimating parameter uncertainty of a DNN, called Bayesian by backpropagation (BBP).
MC dropout~\cite{gal2016dropout} utilized the standard dropout~\cite{hinton2012improving} as an MC sampler to study the dropout uncertainty properties. However, as the aforementioned MC sampling-based methods cannot satisfy the requirement of inference speed~\cite{kong2020sde}, more explicit distributional assumption-based methods have been proposed in recent years to address this problem.

Among UI methods based on explicit distributional assumptions, Hendrycks and Gimpel~\cite{hendrycks2017a} introduced a baseline model that assumes the outputs following softmax distributions and detects the misclassified or the out-of-distribution samples with maximum softmax probabilities in multiple tasks. In~\cite{malinin2018predictive}, the authors presented Dirichlet prior network (DPN) to introduce Dirichlet distributions into the DNNs for modeling the uncertainty. Following the DPN, reverse Kullback-Leibler (RKL) divergence between Dirichlet distributions was introduced for prior network training to improve the UI and adversarial robustness~\cite{malinin2019reverse}. These methods calculate only the means or the modes of the predictions for the UI.

\begin{figure*}[!t]
    \begin{center}
        \includegraphics[width=0.99\linewidth]{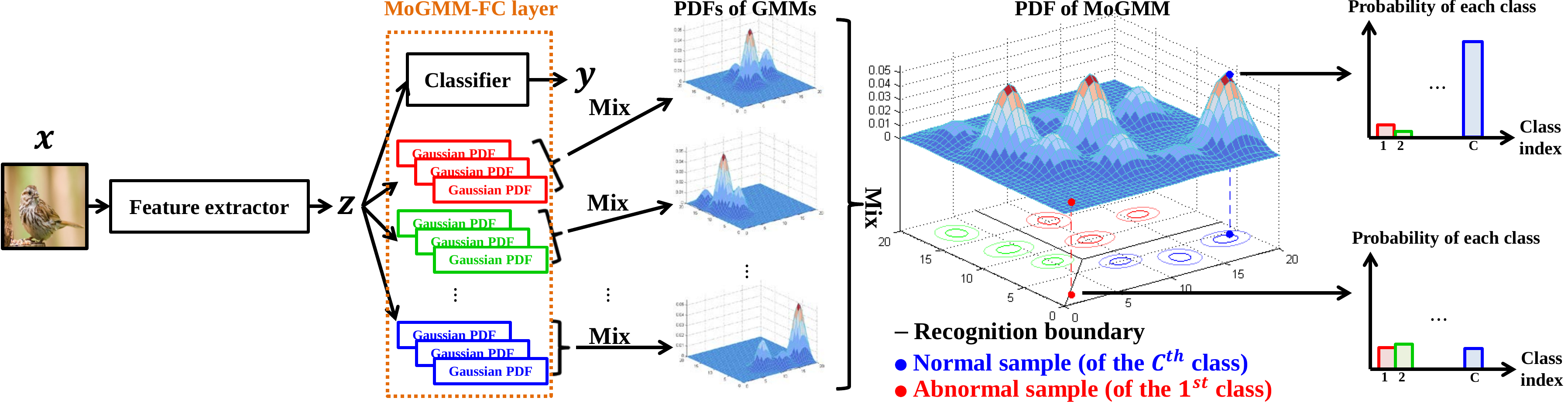}
    \end{center}
    \caption{Structure of the MoGMM-FC layer. The MoGMM is paralleled with the classifier. Both of them are cascaded after the feature extractor. In the MoGMM-FC layer, $C$ GMMs (one GMM for each class) are mixed and their output probabilities are used for the UI. Here, we take three Gaussian components ($K=3$) for each GMM as an example. According to the PDF of the MoGMM, which mixes those of all the GMMs, abnormal samples, including misclassified and out-of-domain/-distribution ones, can be easily detected and distinguished from normal samples, as the probabilities of an abnormal sample belonging to individual classes are all small (bottom-right), unlike the pattern of a normal sample (top-right).}
    \label{fig:structure}
\end{figure*}

In addition to the above conventionally used methods, neural stochastic differential equation (SDE) network (SDE-Net) in~\cite{kong2020sde}, a non-Bayesian method, brought the concept of SDE into the UI. The SDE-Net contains a drift net that controls the system to fit the predictive function and a diffusion net that captures the model uncertainty.

The aforementioned methods only consider the positive samples for the corresponding class centers and jointly train with both the in-domain samples and the out-of-domain samples, which is a close-set out-of-domain detection.

\subsection{Mixture Models and SGVB}\label{ssec:mmandsgvb}

Several works~\cite{piergiovanni2019temporal,variani2015a,wang2018an} have applied GMMs into the DNNs but not for the UI. Variani~\emph{et al.}~\cite{variani2015a} first proposed a GMM layer, which is jointly optimized within a DNN using asynchronous stochastic gradient descent (ASGD). In~\cite{wang2018an}, an unsupervised deep learning framework was proposed to combine deep representations and GMM-based deep modeling. Later on, a temporal Gaussian mixture (TGM) layer was introduced for capturing longer-term temporal information in videos~\cite{piergiovanni2019temporal}. However, no mixtures of mixture models have been explored for jointly modeling the outputs of DNNs or estimating uncertainty.

In addition to some of the aforementioned works, which introduced their own optimization algorithms, various SGVB algorithms~\cite{altosaar2018proximity,fan2015fast,hoffman2013stochastic,plotz2018stochastic,ranganath2014black} have been proposed in recent years for probabilistic model optimization. However, these SGVB algorithms do not consider the negative samples in other classes for the mixture model of a class.

\section{Dual-supervised Uncertainty Inference}\label{sec:methodology}

As the UI usually requires stronger learning ability than other tasks, it is desirable to comprehensively consider both the positive and the negative samples for optimization of each class during training. In this section, we introduce the so-called dual-supervised uncertainty inference (DS-UI) framework to achieve this goal.

Although the classifier of a DNN, commonly the top FC layer, can model the correlation to describe the membership of a feature $\boldsymbol{z}$ belonging to a class, it cannot obtain the uncertainty of $\boldsymbol{z}$ directly. To this end, we propose an MoGMM-FC layer, which can be treated as a probabilistic interpreter for modeling $\boldsymbol{z}$, as shown in Figure~\ref{fig:structure}.
For each GMM in the MoGMM, we propose a dual-supervised SGVB (DS-SGVB) algorithm, which \emph{not only} models the positive samples in the class as the conventional SGVB and the optimization algorithms in other UI methods, \emph{but also} considers the negative samples from other classes. The DS-SGVB can enhance the learning ability of the MoGMM and improve the UI performance by reducing the intra-class distances and enlarging the inter-class margins simultaneously.

\subsection{MoGMM-FC Layer}\label{ssec:mogm}

We propose to use an MoGMM to model the extracted feature vector $\boldsymbol{z}\in R^{M\times1}$, where $M$ is the dimension of $\boldsymbol{z}$. Assuming a recognition task with $C$ classes, we assign a GMM in the MoGMM to each class. The probability density function (PDF) of the MoGMM is defined as
\begin{align}
\begin{footnotesize}
    \text{MoGMM}(\boldsymbol{z};\boldsymbol{\mu},\boldsymbol{\Sigma},\boldsymbol{\eta},\boldsymbol{\omega})=\sum_{i=1}^C\omega_i\underbrace{\sum_{j=1}^K \eta_{ij}\mathcal{N}(\boldsymbol{z};\boldsymbol{\mu}_{ij},\boldsymbol{\Sigma}_{ij})}_{\text{GMM}_i(\boldsymbol{z})},
\end{footnotesize}
\end{align}

\noindent with Gaussian distributions $\mathcal{N}(\boldsymbol{z};\boldsymbol{\mu}_{ij},\boldsymbol{\Sigma}_{ij})$,
where $K$ is the number of components in each GMM and $\boldsymbol{\mu}=\left\{\boldsymbol{\mu}_{ij}\right\}$, $\boldsymbol{\Sigma}=\left\{\boldsymbol{\Sigma}_{ij}\right\}$, and $\boldsymbol{\eta}=\left\{\eta_{ij}\right\}$ are the parameter sets of means, covariances, and mixing weights, respectively. $\boldsymbol{\mu}_{ij}$ ($1\times M$ dimensions), $\boldsymbol{\Sigma}_{ij}$ ($M\times M$ dimensions), and $\eta_{ij}$ are means, covariances, and mixing weight of the $j^{th}$ Gaussian component in the $i^{th}$ GMM. For high-dimensional $\boldsymbol{z}$ in practice, $\boldsymbol{\Sigma}_{ij}$ can be defined as a non-singular diagonal matrix for simplicity, which is employed in this paper. Meanwhile, $\boldsymbol{\omega}=[\omega_1,\cdots,\omega_C]^{\text{T}}$ contains $C$ nonnegative mixing weights of the $C$ GMMs and $\sum_{i=1}^C\omega_i=1$. In the recognition task, $\omega_i$ can be roughly estimated by the proportions of each class in the training set beforehand~\cite{bishop06}.

The distribution of $\mu_{ijm}$ is defined as a Gaussian distribution with mean $a_{ijm}$ and variance $b_{ijm}$, where $a_{ijm}$ and $b_{ijm}$ are elements of their corresponding hyperparameter sets $\boldsymbol{A}=\{a_{ijm}\}$ and $\boldsymbol{B}=\{b_{ijm}\}$, respectively. Meanwhile, $\Sigma_{ijmm}$ follows a Dirac delta distribution $\delta(b_{ijm})$ where the value of the PDF is equal to one if $\Sigma_{ijmm}=b_{ijm}$, zero otherwise. We define the parameter set of the MoGMM as $\boldsymbol{\Phi}=\{\boldsymbol{\mu},\boldsymbol{\Sigma},\boldsymbol{V}\}$, where the latent variable matrix $\boldsymbol{V}$ is a $C\times K$-dimensional matrix and each row $\boldsymbol{v}_{i}$ is a one-hot vector following $p(v_{ij}=1)=\eta_{ij}$, and the hyperparameter set as $\boldsymbol{\theta}=\{\boldsymbol{A},\boldsymbol{B},\boldsymbol{\eta}\}$ for optimization.

Here, the mean parameters in $\boldsymbol{\mu}$ are shared with the classifier. As each row $\boldsymbol{W}_i$ of the parameter matrix $\boldsymbol{W}$ of the classifier is described as a class center, we introduce an approximation of $\boldsymbol{W}$ by $\boldsymbol{\mu}$ to align their dimensions. For the $i^{th}$ GMM (representing the $i^{th}$ class) in the MoGMM, the mean $\boldsymbol{W}_{i}$ of the whole GMM can be approximated as $\boldsymbol{W}_{i}\approx\sum_{j=1}^K\eta_{ij}\boldsymbol{\mu}_{ij}$. Thus, the $i^{th}$ output $y_i$ of the classifier for the $i^{th}$ class can be approximated as
\begin{equation}
    y_i\approx\sum_{j=1}^K\eta_{ij}\boldsymbol{\mu}_{ij}\boldsymbol{z},
\end{equation}
assuming the bias vector of the classifier is removed.

As $\boldsymbol{\eta}_i=[\eta_{i1},\cdots,\eta_{iK}]^{\text{T}}$ is normalized, which is a hard regularization in stochastic gradient-based optimization, we define an alternative $\tilde{\boldsymbol{\eta}}_i\in R^{K\times 1}$ to implicitly optimize $\boldsymbol{\eta}_i$ by $\boldsymbol{\eta}_i=\text{softmax}(\tilde{\boldsymbol{\eta}}_i)$. Similarly, an alternative $\tilde{b}_{ijm}$ is introduced for the positive $b_{ijm}$ by $b_{ijm}=e^{\tilde{b}_{ijm}}$.

\subsection{Optimization for the MoGMM-FC Layer}\label{ssec:sgvb}

\subsubsection{Conventional SGVB}\label{sssec:sgvb}

In variational inference (VI), the common approach~\cite{teye2018bayesian} is to optimize the hyperparameters of a probability model by maximizing the lower bound $L(q_{\boldsymbol{\theta}}(\boldsymbol{\Phi});\boldsymbol{D})$ with the approximated posterior distribution $q_{\boldsymbol{\theta}}(\boldsymbol{\Phi})$, where $\boldsymbol{D}=\{\boldsymbol{Z},\boldsymbol{T}\}$ is the dataset, $\boldsymbol{Z}=\{\boldsymbol{z}_i\}_{i=1}^N$ and $\boldsymbol{T}=\{t_i\}_{i=1}^N$ are the inputs and labels, respectively, and $N$ is the number of samples in $\boldsymbol{D}$. $L(q_{\boldsymbol{\theta}}(\boldsymbol{\Phi});\boldsymbol{D})$, which can be considered as the negative Kullback-Leibler (KL) divergence from $q_{\boldsymbol{\theta}}(\boldsymbol{\Phi})$ to the joint distribution $p(\boldsymbol{D},\boldsymbol{\Phi})$, is defined as
\begin{footnotesize}
\begin{align}
    L(q_{\boldsymbol{\theta}}(\boldsymbol{\Phi});\boldsymbol{D})=&\int q_{\boldsymbol{\theta}}(\boldsymbol{\Phi})\ln\frac{p(\boldsymbol{D},\boldsymbol{\Phi})}{q_{\boldsymbol{\theta}}(\boldsymbol{\Phi})}d\boldsymbol{\Phi}\nonumber\\
    =&\underbrace{\int q_{\boldsymbol{\theta}}(\boldsymbol{\Phi})\ln p(\boldsymbol{D}|\boldsymbol{\Phi})d\boldsymbol{\Phi}}_{L_D(q_{\boldsymbol{\theta}}(\boldsymbol{\Phi}))}-\underbrace{\int q_{\boldsymbol{\theta}}(\boldsymbol{\Phi})\ln\frac{q_{\boldsymbol{\theta}}(\boldsymbol{\Phi})}{p(\boldsymbol{\Phi})}d\boldsymbol{\Phi}}_{D_{\text{KL}}(q_{\boldsymbol{\theta}}(\boldsymbol{\Phi})||p(\boldsymbol{\Phi}))},
\end{align}
\end{footnotesize}

\noindent where the first term $L_D(q_{\boldsymbol{\theta}}(\boldsymbol{\Phi}))$ is the expected log-likelihood and the second term $D_{\text{KL}}(q_{\boldsymbol{\theta}}(\boldsymbol{\Phi})||p(\boldsymbol{\Phi}))$ is the KL divergence from $q_{\boldsymbol{\theta}}(\boldsymbol{\Phi})$ to the prior distribution $p(\boldsymbol{\Phi})$.

For the SGVB algorithm, we usually approximate the expected log-likelihood $L_D(q_{\boldsymbol{\theta}}(\boldsymbol{\Phi}))$ by
\begin{footnotesize}
\begin{align}\label{eq:ld}
    L_D(q_{\boldsymbol{\theta}}(\boldsymbol{\Phi}))&=\frac{1}{N}\sum_{\boldsymbol{z}\in\boldsymbol{Z},t\in\boldsymbol{T}}\text{E}_{q_{\boldsymbol{\theta}}(\boldsymbol{\Phi})}\left[\ln p(\boldsymbol{z},t|\boldsymbol{\Phi})\right]\nonumber\\
    &\approx L_D^{\text{SGVB}}(q_{\boldsymbol{\theta}}(\boldsymbol{\Phi}))\nonumber\\
    &=\frac{1}{B}\sum_{b=1}^B\ln\left(\omega_{t_b}\text{GMM}_{t_b}(\boldsymbol{z}_b)\right),
\end{align}
\end{footnotesize}

\noindent where $B$ is batch size and $t_b$ is the label of the $b^{th}$ sample $\boldsymbol{z}_b$. To be able to use the SGVB, the next step is to consider optimizing $\{-L_D^{\text{SGVB}}(q_{\boldsymbol{\theta}}(\boldsymbol{\Phi}))+\gamma D_{\text{KL}}(q_{\boldsymbol{\theta}}(\boldsymbol{\Phi})||p(\boldsymbol{\Phi}))\}$ with nonnegative multiplier $\gamma$, where the KL divergence is seen as a regularization term. Note that the previous methods~\cite{gal2016dropout,teye2018bayesian} are computationally expensive by making use of the MC estimation approaches. We propose to derive a generalized form for the KL divergence in Section~\ref{sssec:kl} as a regularization term to constrain the hyperparameters in $\boldsymbol{\theta}$.

Note that although the closed-form solution of the MoGMM optimization under the VI framework can be found, it is infeasible to be extended to an SGVB solution, which makes it difficult to jointly optimize the MoGMM together with the classifier.

\begin{table*}[!t]
  \centering
  \caption{Ablation studies with VGG$16$ on the CIFAR-$10$ dataset for misclassification detection. The number of components in each GMM (\emph{i.e.}, $K$) is discussed. The effectiveness of two key parts in the DS-SGVB algorithm,~\emph{i.e.}, $L_D^{\text{NSGVB}}(q_{\boldsymbol{\theta}}(\boldsymbol{\Phi}))$ and $\text{Reg}(q_{\boldsymbol{\theta}}(\boldsymbol{\Phi}))$, are discussed as well. ``$\checkmark$'' means the part is contained and ``$\bigcirc$'' means replacing $\text{Reg}(q_{\boldsymbol{\theta}}(\boldsymbol{\Phi}))$ by the original $D_{\text{KL}}(q_{\boldsymbol{\theta}}(\boldsymbol{\Phi})||p(\boldsymbol{\Phi}))$. The best results are highlighted in~\textbf{bold}.}
  \resizebox{0.895\linewidth}{!}{
    \begin{tabular}{|c|c|c|c|cc|cc|}
    \hline
    \multirow{2}[0]{*}{$L_D^{\text{NSGVB}}(q_{\boldsymbol{\theta}}(\boldsymbol{\Phi}))$} & \multirow{2}[0]{*}{$\text{Reg}(q_{\boldsymbol{\theta}}(\boldsymbol{\Phi}))$} & \multirow{2}[0]{*}{$K$} & \multirow{2}[0]{*}{Accuracy (\%)} & \multicolumn{2}{c|}{AUROC (\%)} & \multicolumn{2}{c|}{AUPR (\%)} \\
          &       &       &       & Max.P. & Ent.  & Max.P. & Ent. \\
    \hline
    \hline
    $\checkmark$ & $\checkmark$ & 1     & $92.09\pm0.15$ & $91.27\pm0.35$ & $91.22\pm0.36$ & $46.14\pm1.23$ & $46.51\pm1.64$ \\
    $\checkmark$ & $\checkmark$ & 2     & $92.12\pm0.14$ & $91.12\pm0.38$ & $91.11\pm0.38$ & $46.01\pm0.83$ & $46.76\pm1.03$ \\
    $\checkmark$ & $\checkmark$ & 4     & $92.16\pm0.09$ & $91.97\pm0.38$ & $91.93\pm0.38$ & $49.76\pm0.88$ & $49.24\pm1.05$ \\
    $\checkmark$ & $\checkmark$ & 8     & $\boldsymbol{92.64\pm0.31}$ & $\boldsymbol{93.51\pm0.27}$ & $\boldsymbol{93.48\pm0.27}$ & $\boldsymbol{53.60\pm0.85}$ & $\boldsymbol{53.25\pm0.48}$ \\
    \hline
    $\checkmark$ & $\bigcirc$ & 8     & $92.28\pm0.76$ & $92.67\pm0.21$ & $92.74\pm0.30$ & $50.05\pm0.53$ & $50.30\pm0.27$ \\
    $\checkmark$ &           & 8     & $92.36\pm0.25$ & $90.89\pm0.47$ & $90.88\pm0.49$ & $46.69\pm2.30$ & $47.24\pm2.64$ \\
              & $\checkmark$ & 8     & $92.57\pm0.10$ & $91.37\pm0.26$ & $91.32\pm0.28$ & $45.95\pm0.56$ & $46.81\pm1.08$ \\
              &           & 8     & $92.36\pm0.08$ & $91.09\pm0.34$ & $91.03\pm0.36$ & $45.52\pm0.88$ & $46.37\pm0.93$ \\
    \hline
    \end{tabular}}
  \label{tab:ablation}
\end{table*}

\subsubsection{Dual-supervised SGVB}\label{sssec:dssgvb}

In this section, we propose the DS-SGVB algorithm to reduce the intra-class distances and enlarge the inter-class margins simultaneously. Recall that the approximated expected log-likelihood in~\eqref{eq:ld} undertakes ``pull'' operation between the class centers and their corresponding positive samples, we define a dual-supervised expected log-likelihood $L_D^{\text{DS}}(q_{\boldsymbol{\theta}}(\boldsymbol{\Phi}))$ as
\begin{equation}\label{eq:ld2}
    L_D^{\text{DS}}(q_{\boldsymbol{\theta}}(\boldsymbol{\Phi}))=L_D^{\text{SGVB}}(q_{\boldsymbol{\theta}}(\boldsymbol{\Phi}))-\rho L_D^{\text{NSGVB}}(q_{\boldsymbol{\theta}}(\boldsymbol{\Phi})),
\end{equation}

\noindent where $\rho$ is a nonnegative multiplier and $L_D^{\text{NSGVB}}(q(\boldsymbol{\Phi}))$ is the negative-sample expected log-likelihood as
\begin{equation}
    L_D^{\text{NSGVB}}(q_{\boldsymbol{\theta}}(\boldsymbol{\Phi}))=\frac{1}{B}\sum_{b=1}^B\sum_{i\neq t_b}\ln \left(\omega_i\text{GMM}_i(\boldsymbol{z}_b)\right),
\end{equation}
which minimizes the log-likelihood of each GMM \emph{w.r.t.} negative samples and undertakes ``push'' operation between the class centers and the negative samples belonging to other classes. By minimizing $-L_D^{\text{DS}}(q_{\boldsymbol{\theta}}(\boldsymbol{\Phi}))$, the learning ability of the MoGMM can be further enhanced, as it not only models the positive samples in the class for a GMM as the conventional SGVB, but also considers the negative samples from other classes.

\subsubsection{Generalized Form of $D_{\text{KL}}(q_{\boldsymbol{\theta}}(\boldsymbol{\Phi})||p(\boldsymbol{\Phi}))$}\label{sssec:kl}

In this section, a regularization term $\text{Reg}(q_{\boldsymbol{\theta}}(\boldsymbol{\Phi}))$, which is related to $D_{\text{KL}}(q_{\boldsymbol{\theta}}(\boldsymbol{\Phi})||p(\boldsymbol{\Phi}))$ and performs as a generalized form of it, is applied to constrain the hyperparameters in $\boldsymbol{\theta}$ of the MoGMM.

\noindent \textbf{Proposition 1.} \textit{Let the prior distributions of $\mu_{ijm}$, $\Sigma_{ijmm}$, and $\boldsymbol{v}_{i}$ be standard normal distribution, uniform distribution in the interval of $(0,\infty)$ and categorical distribution with equal probabilities, respectively. The generalized form $\text{Reg}(q_{\boldsymbol{\theta}}(\boldsymbol{\Phi}))$ of $D_{\text{KL}}(q_{\boldsymbol{\theta}}(\boldsymbol{\Phi})||p(\boldsymbol{\Phi}))$ is}

\begin{footnotesize}
\begin{align}
    &\text{Reg}(q_{\boldsymbol{\theta}}(\boldsymbol{\Phi}))\nonumber\\
    &=\sum_{i=1}^C\omega_i^*\sum_{j=1}^K\eta_{ij}^*\sum_{m=1}^M D_{\text{KL}}(q(\mu_{ijm}|a_{ijm},b_{ijm})||p(\mu_{ijm}))\nonumber\\
    &+\sum_{i=1}^C\omega_i^*\sum_{j=1}^K\eta_{ij}^*\sum_{m=1}^M D_{\text{KL}}(q(\Sigma_{ijmm}|b_{ijm})||p(\Sigma_{ijmm}))\nonumber\\
    &+\sum_{i=1}^C\omega_i^* D_{\text{KL}}(q(\boldsymbol{v}_i|\boldsymbol{\eta}_i)||p(\boldsymbol{v}_i))\nonumber\\
    &=\sum_{i=1}^C\omega_i^*\left\{\sum_{j=1}^K\left[\vphantom{\sum_k}\eta_{ij}\ln(\eta_{ij}\cdot K)\right.\right.\nonumber\\
    &+\left.\left.\frac{\eta_{ij}^*}{2}\vphantom{\sum_{m=1}^M}\sum_{m=1}^M\left(b_{ijm}+a_{ijm}^2-\ln b_{ijm}-1\right) \vphantom{\sum_k}\right]\vphantom{\sum_j^M}\right\},
\end{align}
\end{footnotesize}

\noindent \textit{where $\boldsymbol{\Sigma}_{ij}$ is assumed to be a non-singular diagonal matrix. $\omega_i^*$ and $\eta_{ij}^*$ are nonnegative sub-multipliers and set equal to $\omega_i$ and $\eta_{ij}$, respectively, in this paper. Note that $\text{Reg}(q_{\boldsymbol{\theta}}(\boldsymbol{\Phi}))$ is equivalent to the original $D_{\text{KL}}(q_{\boldsymbol{\theta}}(\boldsymbol{\Phi})||p(\boldsymbol{\Phi}))$ when $\omega_i^*$ and $\eta_{ij}^*$ are equal to one.}

Compared with $D_{\text{KL}}(q_{\boldsymbol{\theta}}(\boldsymbol{\Phi})||p(\boldsymbol{\Phi}))$, the superiority of $\text{Reg}(q_{\boldsymbol{\theta}}(\boldsymbol{\Phi}))$ is that we can adaptively optimize $\eta_{ij}^*$ for the KL divergences of $\mu_{ijm}$ and $\Sigma_{ijmm}$ during training, rather than setting them as equal weights. In addition, the sub-multipliers in $\text{Reg}(q_{\boldsymbol{\theta}}(\boldsymbol{\Phi}))$ can also reflect the contributions of the KL divergences of $\mu_{ijm}$ and $\Sigma_{ijmm}$ in the MoGMM optimization.

In the end, the total loss function $\mathcal{L}$ is defined as
\begin{equation}\label{eq:loss}
    \mathcal{L}=L_{\text{CE}}-L_D^{\text{DS}}(q_{\boldsymbol{\theta}}(\boldsymbol{\Phi})) +\gamma\text{Reg}(q_{\boldsymbol{\theta}}(\boldsymbol{\Phi})),
\end{equation}

\noindent where $L_{\text{CE}}$ is the cross-entropy (CE) loss for the classifier in Figure~\ref{fig:structure} and $\gamma$ is a nonnegative multiplier.

\section{Experimental Results and Discussions}\label{sec:experimentalresults}

We conducted three different UI tasks, including the misclassification detection, the open-set out-of-domain detection, and the open-set out-of-distribution detection. The proposed DS-UI was evaluated with VGG$16$~\cite{simonyan2015very} and ResNet$18$~\cite{he2016deep} as backbone models on CIFAR-$10$/-$100$~\cite{krizhevsky09cifar}, street view house numbers (SVHN)~\cite{netzer2011reading}, and tiny ImageNet (TIM)~\cite{russakovsky2015imagenet} datasets. We compared the DS-UI with the baseline~\cite{hendrycks2017a}, the MC dropout~\cite{gal2016dropout}, the DPN~\cite{malinin2018predictive}, the RKL~\cite{malinin2019reverse}, and the SDE-Net~\cite{kong2020sde}. In addition to the UI methods, we also compared the DS-UI with some classic open-set recognition methods, G-OpenMax~\cite{ge2017generative}, C$2$AE~\cite{Oza2019c2ae}, and GDOSR~\cite{perera2020generative}, for the open-set out-of-domain detection.

\begin{table*}[!t]
  \centering
  \caption{Means and standard deviations of image recognition accuracies (\%) on the four datasets. Note that ``-'' means that the model do not work in the case of reimplementation, ``$\checkmark$'' means statistically significant difference between the accuracies of the DS-UI and those of the referred methods, ``$\times$'' means no significance, and ``N/A'' means inapplicable. The best results are highlighted in~\textbf{bold}.}
  \resizebox{\linewidth}{!}{
    \begin{tabular}{|l|@{}c@{}c@{}|@{}c@{}c@{}|@{}c@{}c@{}|@{}c@{}c@{}|}
    \hline
    \multicolumn{1}{|c|}{Dataset} & \multicolumn{2}{c|}{CIFAR-$10$} & \multicolumn{2}{c|}{CIFAR-$100$} & \multicolumn{2}{c|}{SVHN} & \multicolumn{2}{c|}{TIM} \\
    \hline
    \multicolumn{1}{|c|}{Method} & VGG$16$ & ResNet$18$ & VGG$16$ & ResNet$18$ & VGG$16$ & ResNet$18$ & VGG$16$ & ResNet$18$ \\
    \hline
    \hline
    Baseline (ICLR$2017$) & $91.76\pm0.09$\ {\tiny($\checkmark$)} & $92.56\pm0.14$\ {\tiny($\checkmark$)} & $70.46\pm0.24$\ {\tiny($\checkmark$)} & $71.04\pm0.19$\ {\tiny($\checkmark$)} & $95.25\pm0.11$\ {\tiny($\checkmark$)} & $95.23\pm0.13$\ {\tiny($\checkmark$)} & $46.13\pm0.41$\ {\tiny($\checkmark$)} & $50.37\pm0.45$\ {\tiny($\checkmark$)} \\
    MC dropout (ICML$2016$) & $91.76\pm0.09$\ {\tiny($\checkmark$)} & $92.56\pm0.14$\ {\tiny($\checkmark$)} & $70.46\pm0.24$\ {\tiny($\checkmark$)} & $71.04\pm0.19$\ {\tiny($\checkmark$)} & $95.25\pm0.11$\ {\tiny($\checkmark$)} & $95.23\pm0.13$\ {\tiny($\checkmark$)} & $46.13\pm0.41$\ {\tiny($\checkmark$)} & $50.37\pm0.45$\ {\tiny($\checkmark$)} \\
    DPN (NeurIPS$2018$)   & $90.73\pm0.35$\ {\tiny($\checkmark$)} & $91.98\pm0.32$\ {\tiny($\checkmark$)} & $67.56\pm0.28$\ {\tiny($\checkmark$)} & $69.80\pm0.11$\ {\tiny($\checkmark$)} & $93.51\pm0.42$\ {\tiny($\checkmark$)} & $93.98\pm0.67$\ {\tiny($\checkmark$)} & -     & - \\
    RKL (NeurIPS$2019$)   & $92.25\pm0.30$\ {\tiny($\times$)} & $92.37\pm0.23$\ {\tiny($\checkmark$)} & $70.58\pm0.26$\ {\tiny($\checkmark$)} & $71.62\pm0.76$\ {\tiny($\times$)} & $94.95\pm0.07$\ {\tiny($\checkmark$)} & $95.09\pm0.09$\ {\tiny($\checkmark$)} & $45.21\pm0.16$\ {\tiny($\checkmark$)} & $50.31\pm0.29$\ {\tiny($\checkmark$)} \\
    SDE-Net (ICML$2020$) & -     & $92.15\pm0.76$\ {\tiny($\checkmark$)} & -     & $52.52\pm1.80$\ {\tiny($\checkmark$)} & -     & $95.00\pm0.23$\ {\tiny($\checkmark$)} & -     & - \\
    DS-UI (Ours) & \ \ $\boldsymbol{92.64\pm0.31}$\ {\tiny(\text{N/A})} \ \ & \ \ $\boldsymbol{93.09\pm0.04}$\ {\tiny(\text{N/A})} \ \ & \ \ $\boldsymbol{71.39\pm0.38}$\ {\tiny(\text{N/A})} \ \ & \ \ $\boldsymbol{71.94\pm0.43}$\ {\tiny(\text{N/A})} \ \ & \ \ $\boldsymbol{95.71\pm0.11}$\ {\tiny(\text{N/A})} \ \ & \ \ $\boldsymbol{95.94\pm0.11}$\ {\tiny(\text{N/A})} \ \ & \ \ $\boldsymbol{46.83\pm0.28}$\ {\tiny(\text{N/A})} \ \ & \ \ $\boldsymbol{52.02\pm0.30}$\ {\tiny(\text{N/A})} \ \ \\
    \hline
    \end{tabular}}
  \label{tab:cls}
\end{table*}

\begin{figure*}[!t]
    \centering
    \resizebox{1.01\linewidth}{!}{
    \begin{tabular}{l}
    \includegraphics[width=0.92\linewidth]{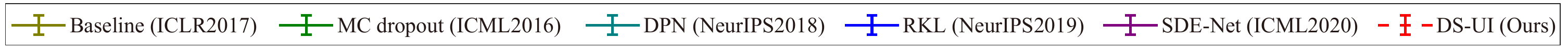} \\
    \begin{subfigure}[t]{0.249\linewidth}
        \centering
        \includegraphics[width=1\linewidth]{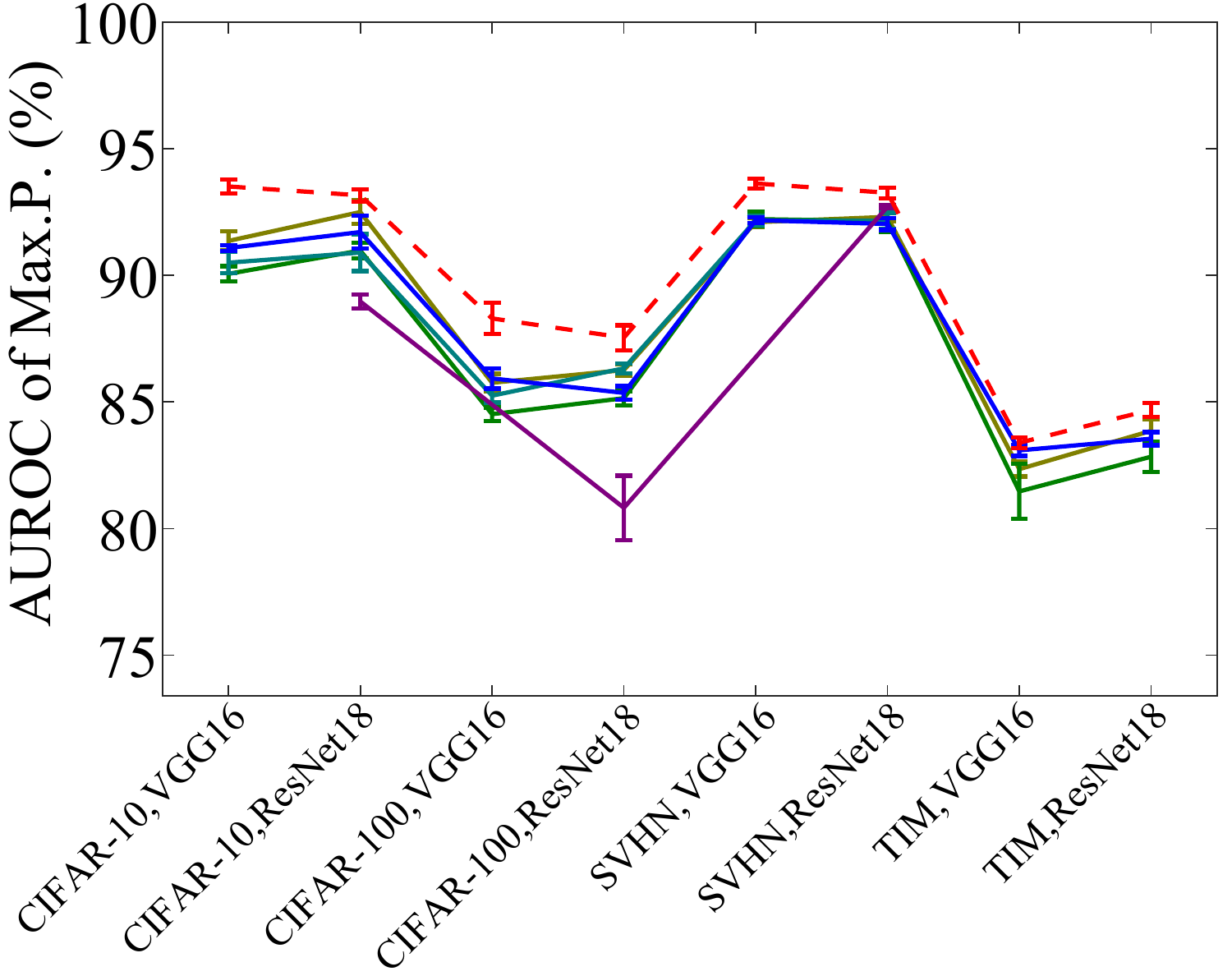}
        \subcaption{AUROC (\%) of Max.P.}
    \end{subfigure}
    \begin{subfigure}[t]{0.249\linewidth}
        \centering
        \includegraphics[width=1\linewidth]{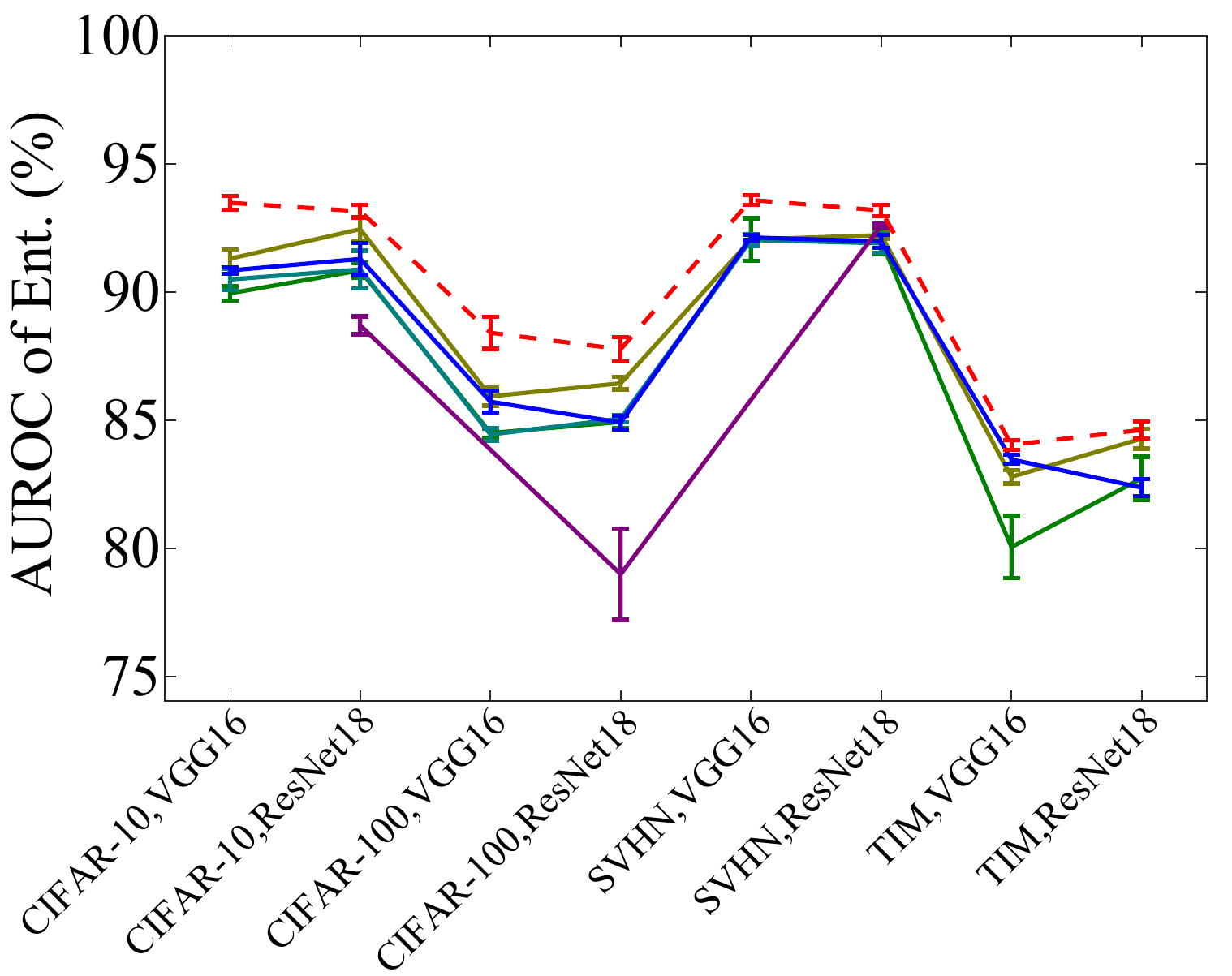}
        \subcaption{AUROC (\%) of Ent.}
    \end{subfigure}
    \begin{subfigure}[t]{0.249\linewidth}
        \centering
        \includegraphics[width=1\linewidth]{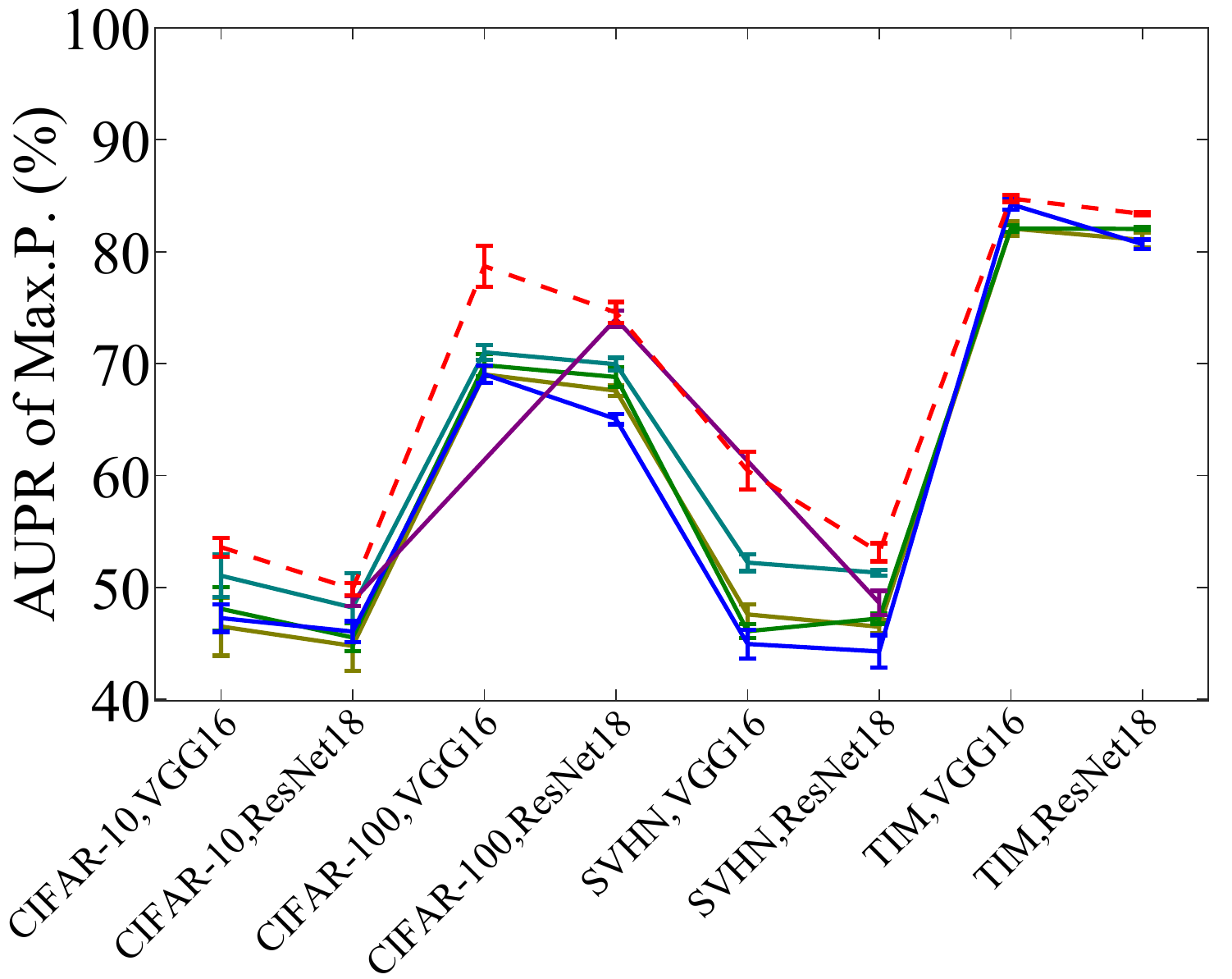}
        \subcaption{AUPR (\%) of Max.P.}
    \end{subfigure}
    \begin{subfigure}[t]{0.249\linewidth}
        \centering
        \includegraphics[width=1\linewidth]{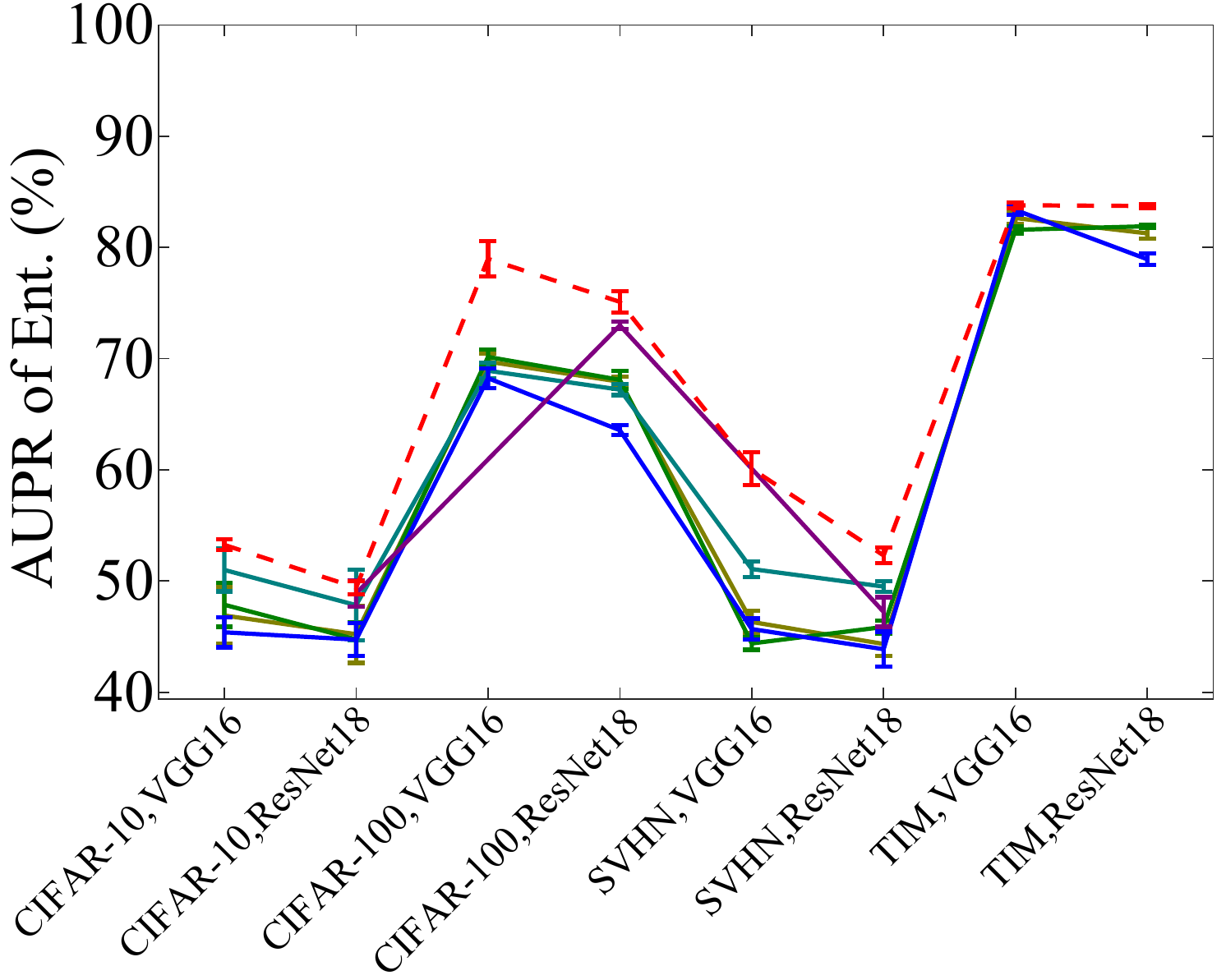}
        \subcaption{AUPR (\%) of Ent.}
    \end{subfigure}
    \end{tabular}}
    \caption{Performance of misclassification detection with the two backbones on the four datasets. Note that annotations in x-axis mean ``dataset, backbone''. The error bars represent standard deviations of the values of the metrics for the methods. The dashed lines in each subfigure present the DS-UI and the other solid lines present the referred methods.}\label{fig:miscls}
\end{figure*}

\begin{figure}[!t]
    \centering
    \begin{subfigure}[t]{\linewidth}
        \centering
        \includegraphics[width=1\linewidth]{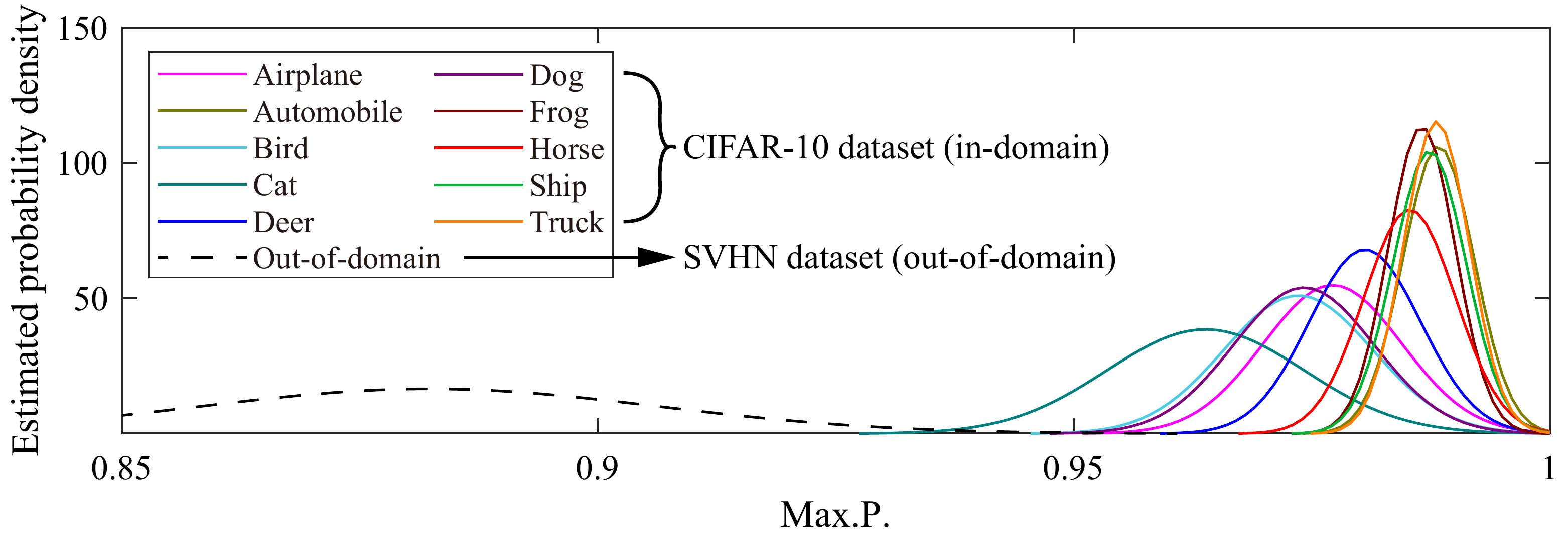}
        \subcaption{Distributions of Max.P.}
    \end{subfigure}
    \begin{subfigure}[t]{\linewidth}
        \centering
        \includegraphics[width=1\linewidth]{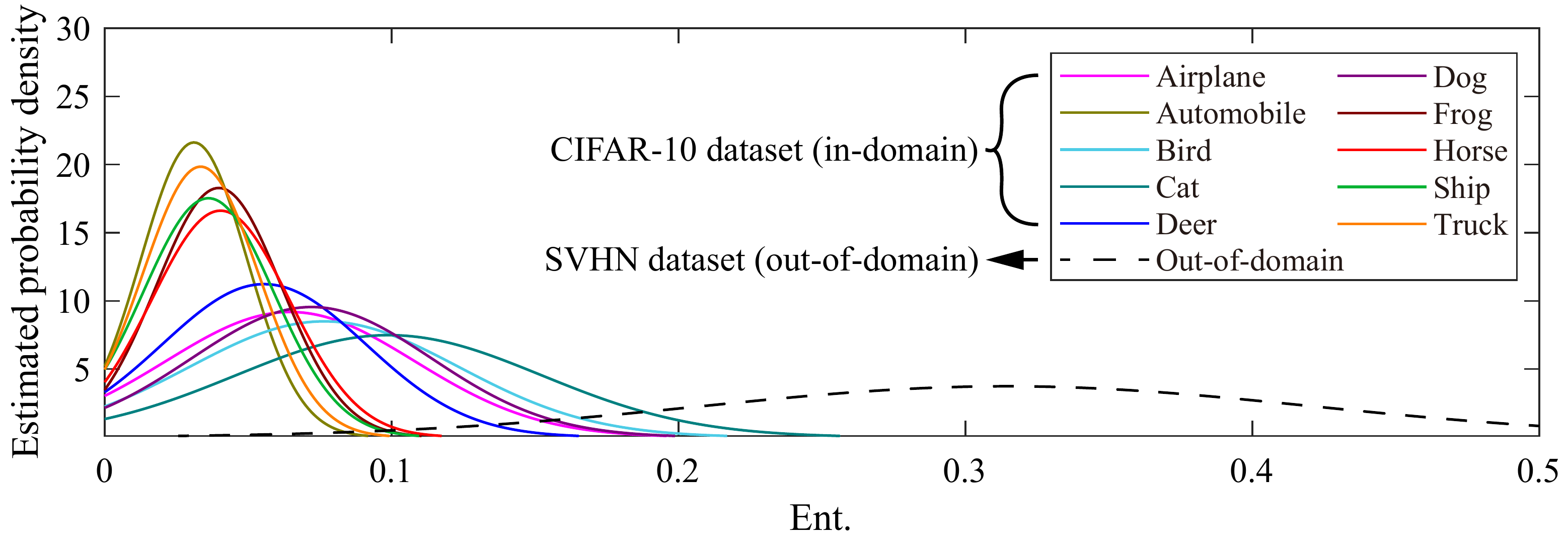}
        \subcaption{Distributions of Ent.}
    \end{subfigure}
    \caption{Estimated Gaussian distributions of Max.P. and Ent. values in the open-set out-of-domain detection. Test sets of the ``CIFAR-$10\to$SVHN'' pair are chosen.}\label{fig:metricdistrib}
\end{figure}

\subsection{Implementation Details}\label{ssec:implementation}

Following the settings in~\cite{hendrycks2017a,malinin2018predictive}, we introduced max probability (Max.P.) and entropy (Ent.) of output probabilities as uncertainty measurement and adopted the area under receiver operating characteristic curve (AUROC) and the area under precision-recall curve (AUPR) for evaluations. The AUROC and the AUPR are the larger the better.

In model training, we applied Adam~\cite{kingma2015adam} optimizer with $100$ epochs for CIFAR-$10$/-$100$, $40$ epochs for SVHN, and $120$ epochs for TIM. We used $1$-cycle learning rate scheme, where we set initial learning rates as $7.5\times10^{-4}$ for each dataset and cycle length as $70$ epochs for CIFAR-$10$/-$100$, $30$ epochs for SVHN, and $80$ epochs for TIM. Weight decay values were set as $5\times10^{-4}$. $\gamma$ and $\rho$ were set as $1\times10^{-4}$ and $4$, respectively. We performed the same training strategy to the referred methods. Following~\cite{hendrycks2017a,malinin2018predictive}, the FC layers of VGG$16$ and ResNet$18$ are replaced by a three-layer FC net with $2048$ hidden units for each hidden layer. Leaky ReLU~\cite{maas2013rectifier} was used as the activation function. Hyperparameters of the referred methods were set the same as those in the original papers.

For all the methods, we conducted five runs and report the means and the standard deviations of recognition accuracies, the AUROC and the AUPR. The SDE-Net can be implemented with the ResNet structure only and the DPN does not work on the TIM dataset in practice. Please find other details of implementation in the supplementary material. We conducted unpaired Student's~\emph{t}-tests between the values of the metrics with significance level as $0.05$.

\begin{figure*}[!t]
    \centering
    \resizebox{1.01\linewidth}{!}{
    \begin{tabular}{l}
    \includegraphics[width=0.92\linewidth]{performance_caption-v3.pdf} \\
    \begin{subfigure}[t]{0.249\linewidth}
        \centering
        \includegraphics[width=1\linewidth]{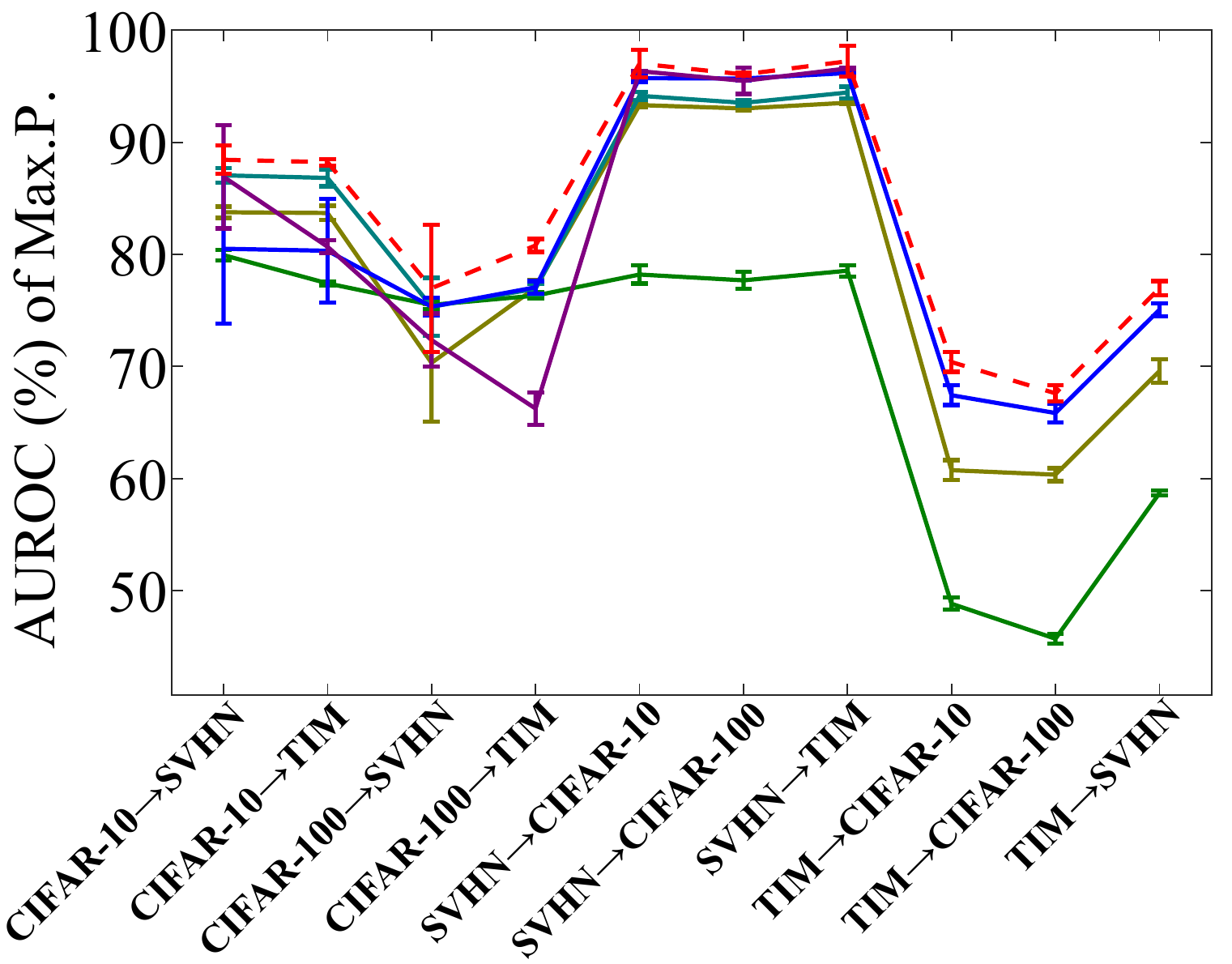}
        \subcaption{AUROC (\%) of Max.P.}
    \end{subfigure}
    \begin{subfigure}[t]{0.249\linewidth}
        \centering
        \includegraphics[width=1\linewidth]{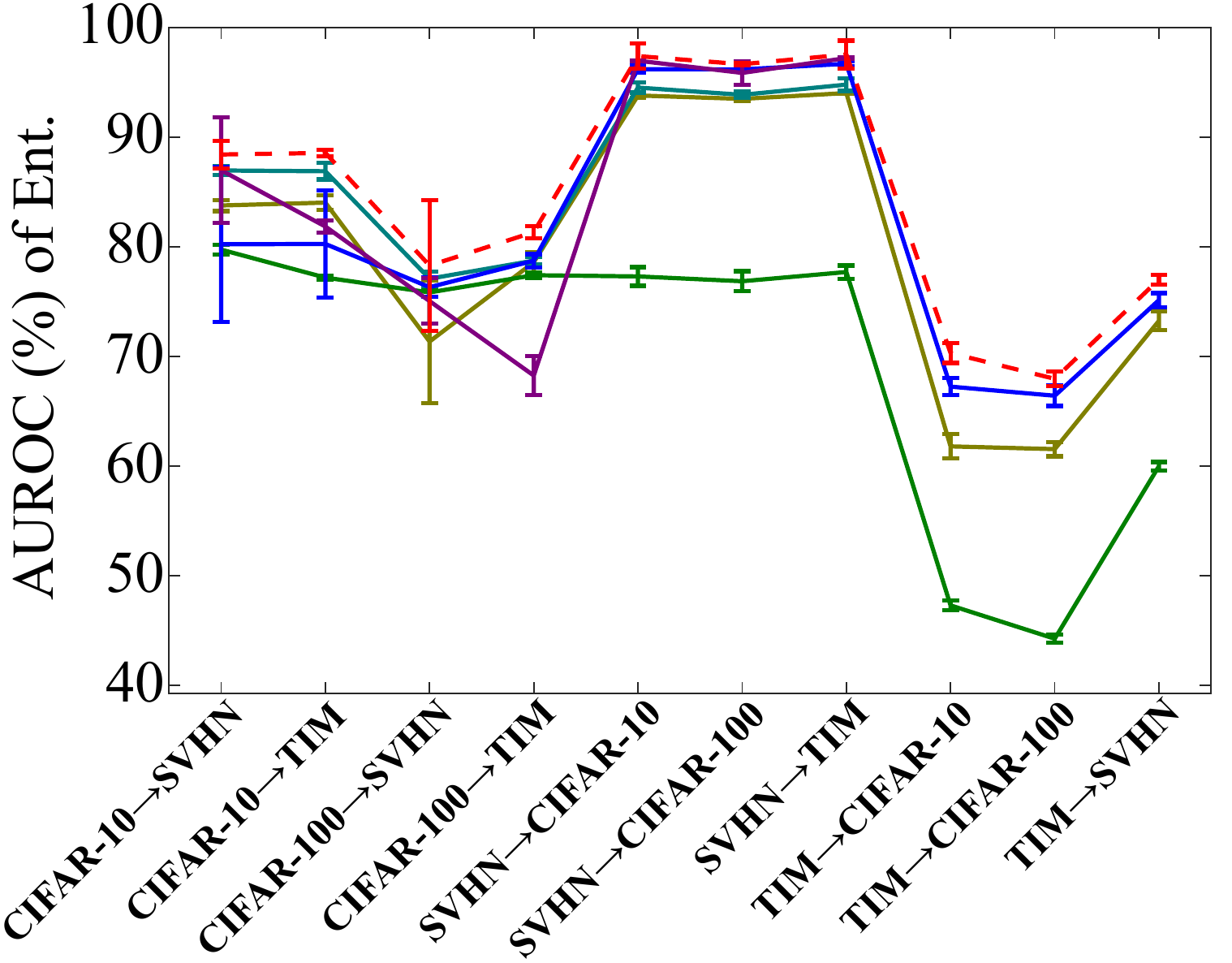}
        \subcaption{AUROC (\%) of Ent.}
    \end{subfigure}
    \begin{subfigure}[t]{0.249\linewidth}
        \centering
        \includegraphics[width=1\linewidth]{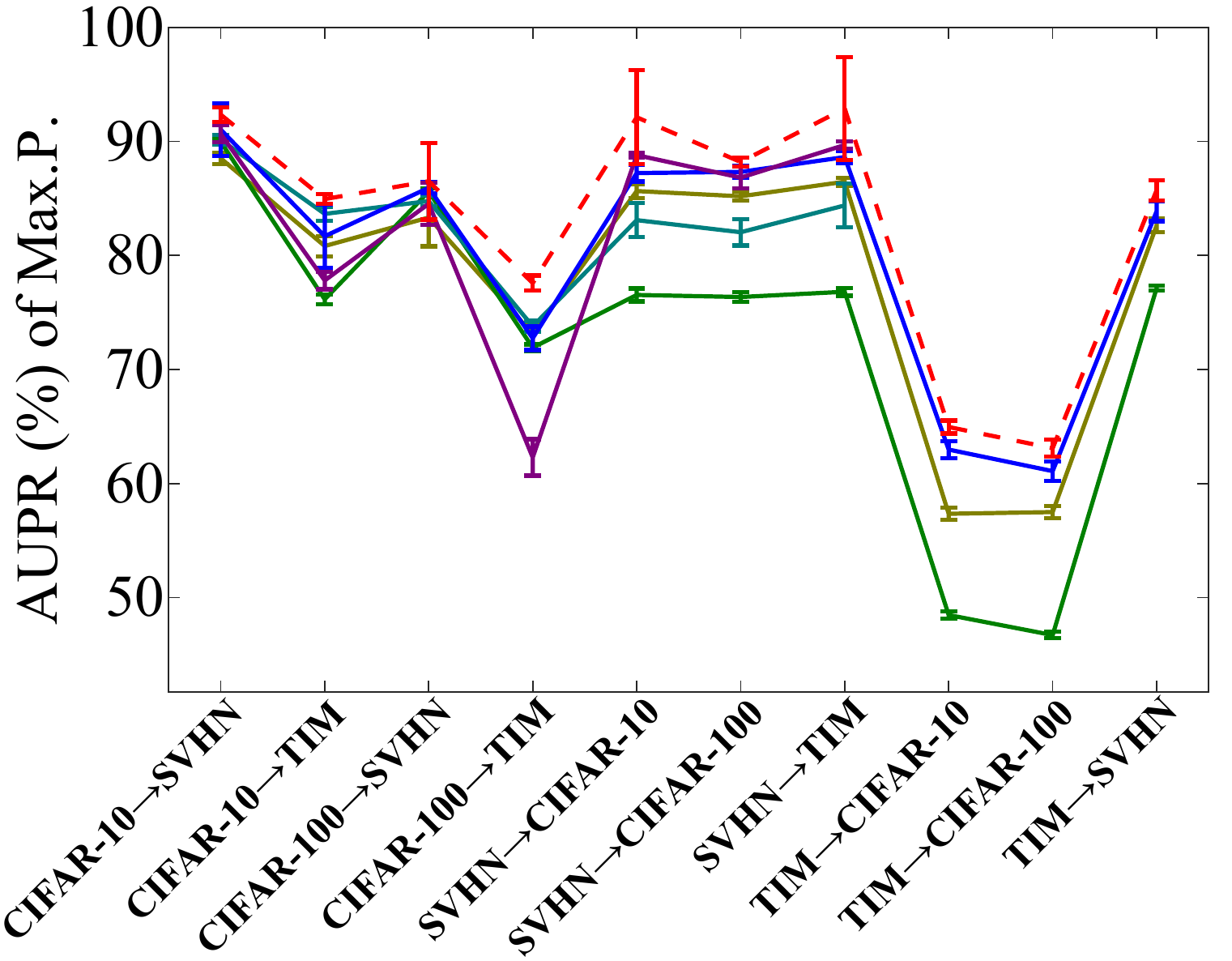}
        \subcaption{AUPR (\%) of Max.P.}
    \end{subfigure}
    \begin{subfigure}[t]{0.249\linewidth}
        \centering
        \includegraphics[width=1\linewidth]{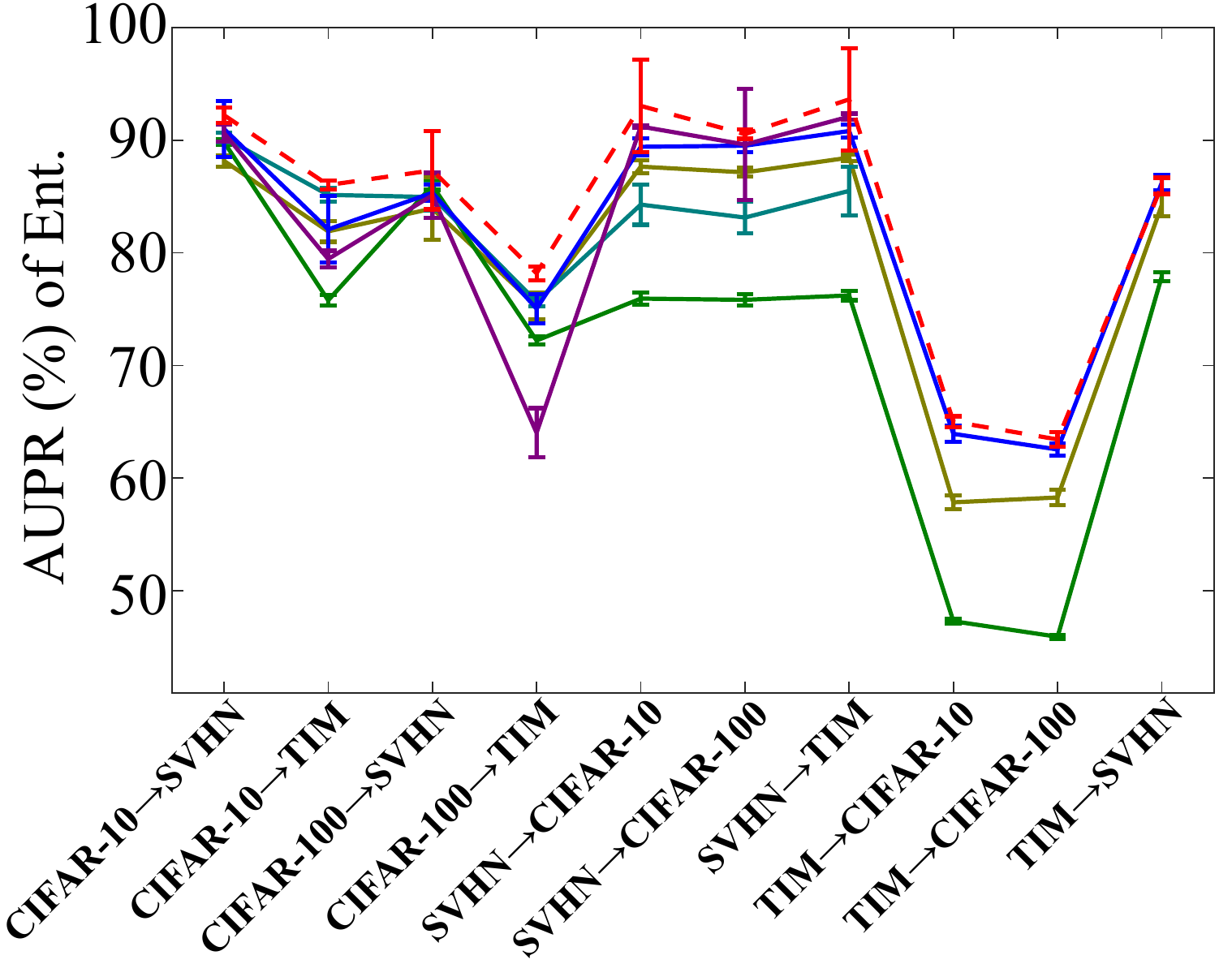}
        \subcaption{AUPR (\%) of Ent.}
    \end{subfigure}
    \end{tabular}}
    \caption{Performance of open-set out-of-domain detection with ResNet$18$ on ten dataset pairs. Note that annotations in x-axis mean ``in-domain dataset$\to$out-of-domain dataset''. The error bars represent standard deviations of the values of the metrics for the methods. The dashed lines in each subfigure present the DS-UI and the solid lines present the referred ones.}\label{fig:ood}
\end{figure*}

\begin{table*}[!t]
  \centering
  \caption{Performance of open set out-of-domain detection on the CIFAR-$10$ and the TIM datasets (another setting). The datasets are divided into in-domain (ID) and out-of-domain (OoD) sets, respectively. ``\#ID'' and ``\#OoD'' mean the class numbers of the ID set and the OoD set, respectively. Note that ``${\dagger}$'' means the results in the row are obtained from~\cite{perera2020generative}, ``$\checkmark$'' means statistically significant difference between the values of the evaluation metrics of the DS-UI and those of the referred methods, ``$\times$'' means no significance, and ``N/A'' means inapplicable. The best results in each case are highlighted in~\textbf{bold}.}
  \resizebox{\linewidth}{!}{
    \begin{tabular}{|l|cc|cc|cc|cc|}
    \hline
    \multicolumn{1}{|c|}{Dataset} & \multicolumn{4}{c|}{CIFAR-$10$ (\#ID: $6$, \#OoD: $4$)} & \multicolumn{4}{c|}{TIM (\#ID: $20$, \#OoD: $180$)} \\
    \hline
    \multicolumn{1}{|c|}{Metric} & \multicolumn{2}{c|}{AUROC (\%)} & \multicolumn{2}{c|}{AUPR (\%)} & \multicolumn{2}{c|}{AUROC (\%)} & \multicolumn{2}{c|}{AUPR (\%)} \\
    \hline
    \multicolumn{1}{|c|}{Method} & Max.P. & Ent.  & Max.P. & Ent.  & Max.P. & Ent.  & Max.P. & Ent. \\
    \hline
    \hline
    MC dropout (ICML$2016$) & $66.88\pm0.28$\ {\tiny($\checkmark$)} & $66.61\pm0.27$\ {\tiny($\checkmark$)} & $54.89\pm0.45$\ {\tiny($\checkmark$)} & $54.22\pm0.44$\ {\tiny($\checkmark$)} & $65.09\pm0.37$\ {\tiny($\checkmark$)} & $64.60\pm0.47$\ {\tiny($\checkmark$)} & $93.28\pm0.10$\ {\tiny($\checkmark$)} & $92.98\pm0.11$\ {\tiny($\checkmark$)} \\
    RKL (NeurIPS$2019$)  & $76.47\pm0.94$\ {\tiny($\checkmark$)} & $76.58\pm0.99$\ {\tiny($\checkmark$)} & $65.49\pm0.52$\ {\tiny($\times$)} & $66.42\pm0.49$\ {\tiny($\times$)} & $70.30\pm0.80$\ {\tiny($\checkmark$)} & $70.89\pm0.82$\ {\tiny($\checkmark$)} & $93.85\pm0.17$\ {\tiny($\checkmark$)} & $93.97\pm0.13$\ {\tiny($\checkmark$)} \\
    SDE-Net (ICML$2020$) & $77.02\pm0.81$\ {\tiny($\checkmark$)} & $77.94\pm0.79$\ {\tiny($\checkmark$)} & $64.57\pm0.85$\ {\tiny($\checkmark$)} & $66.53\pm0.82$\ {\tiny($\times$)} & $65.68\pm0.99$\ {\tiny($\checkmark$)} & $66.95\pm1.06$\ {\tiny($\checkmark$)} & $93.51\pm0.31$\ {\tiny($\checkmark$)} & $93.71\pm0.31$\ {\tiny($\checkmark$)} \\
    G-OpenMax (BMVC$2017$)$^{\dagger}$ & $67.50\pm3.50$\ {\tiny($\checkmark$)} & -     & -     & -     & $58.00\pm\text{N/A}$\ {\tiny($\checkmark$)} & -     & -     & - \\
    C$2$AE (CVPR$2019$)$^{\dagger}$ & $71.10\pm0.80$\ {\tiny($\checkmark$)} & -     & -     & -     & $58.10\pm1.90$\ {\tiny($\checkmark$)} & -     & -     & - \\
    GDOSR (CVPR$2020$)$^{\dagger}$ & $80.70\pm3.90$\ {\tiny($\times$)} & -     & -     & -     & $60.80\pm1.70$\ {\tiny($\checkmark$)} & -     & -     & - \\
    DS-UI (Ours) & $\boldsymbol{81.02\pm0.54}$\ {\tiny(\text{N/A})} & $\boldsymbol{81.34\pm0.55}$\ {\tiny(\text{N/A})} & $\boldsymbol{66.26\pm0.88}$\ {\tiny(\text{N/A})} & $\boldsymbol{67.26\pm0.86}$\ {\tiny(\text{N/A})} & $\boldsymbol{72.27\pm0.11}$\ {\tiny(\text{N/A})} & $\boldsymbol{73.10\pm0.24}$\ {\tiny(\text{N/A})} & $\boldsymbol{94.90\pm0.08}$\ {\tiny(\text{N/A})} & $\boldsymbol{95.11\pm0.09}$\ {\tiny(\text{N/A})} \\
    \hline
    \end{tabular}}
  \label{tab:ood2}
\end{table*}

\subsection{Ablation Studies}\label{ssec:ablation}

We conducted ablation studies with VGG$16$ on the CIFAR-$10$ dataset under misclassification detection (Table~\ref{tab:ablation}) to discuss the selection of number of components $K$ in each GMM of the MoGMM, as well as the effectiveness of two key parts in the DS-SGVB,~\emph{i.e.}, $L_D^{\text{NSGVB}}(q_{\boldsymbol{\theta}}(\boldsymbol{\Phi}))$ and $\text{Reg}(q_{\boldsymbol{\theta}}(\boldsymbol{\Phi}))$. Although the accuracies maintain steady in different cases, the AUROC and the AUPR change sharply in the full DS-SGVB after increasing $K$ to eight. Thus, we set $K$ as eight in the following experiments. In addition, the results using $\text{Reg}(q_{\boldsymbol{\theta}}(\boldsymbol{\Phi}))$ surpasses those using $D_{\text{KL}}(q_{\boldsymbol{\theta}}(\boldsymbol{\Phi})||p(\boldsymbol{\Phi}))$. Meanwhile, the AUROC and the AUPR of the full DS-SGVB can outperform those of removing one or two key parts, which means the two key parts are essential and should be combined in implementation.

\begin{table}[!t]
  \centering
  \caption{Performance of open-set out-of-distribution detection between CIFAR-$100$ dataset and the synthetic uniform noise dataset. Means and standard deviations of the metrics (\%) are shown. Note that ``$\checkmark$'' means statistically significant difference between the values of the metrics of the DS-UI and those of the referred methods, ``$\times$'' means no significance, and ``N/A'' means inapplicable. The best and the second best results are highlighted in~\textbf{bold} and~\underline{underline}.}
  \resizebox{1.02\linewidth}{!}{
    \begin{tabular}{|@{}l@{}|@{}c@{}c@{}|@{}c@{}c@{}|}
    \hline
    \multicolumn{1}{|c|}{\multirow{2}[0]{*}{Method}} & \multicolumn{2}{c|}{AUROC (\%)} & \multicolumn{2}{c|}{AUPR (\%)} \\
          & \multicolumn{1}{c}{Max.P.} & \multicolumn{1}{c|}{Ent.} & \multicolumn{1}{c}{Max.P.} & \multicolumn{1}{c|}{Ent.} \\
    \hline
    \hline
    \ \ Baseline (ICLR$2017$) & $84.20\pm4.56$\ {\tiny($\checkmark$)} & $86.70\pm5.75$\ {\tiny($\checkmark$)} & $74.14\pm6.97$\ {\tiny($\checkmark$)} & $77.18\pm8.40$\ {\tiny($\checkmark$)} \\
    \ \ MC dropout (ICML$2016$)\ \ \ & $82.39\pm0.16$\ {\tiny($\checkmark$)} & $84.11\pm0.18$\ {\tiny($\checkmark$)} & $78.24\pm0.31$\ {\tiny($\checkmark$)} & $79.07\pm0.31$\ {\tiny($\checkmark$)} \\
    \ \ DPN (NeurIPS$2018$)  & $88.99\pm3.01$\ {\tiny($\checkmark$)} & $90.35\pm0.64$\ {\tiny($\checkmark$)} & $82.67\pm1.47$\ {\tiny($\checkmark$)} & $83.55\pm2.87$\ {\tiny($\checkmark$)} \\
    \ \ RKL (NeurIPS$2019$)  &\ \ $87.64\pm7.15$\ {\tiny($\checkmark$)}\ \ &\ \ $88.61\pm6.03$\ {\tiny($\checkmark$)}\ \ &\ \ $79.44\pm10.57$\ {\tiny($\checkmark$)}\ \ &\ \ $80.17\pm8.89$\ {\tiny($\checkmark$)}\ \ \\
    \ \ SDE-Net (ICML$2020$)\tablefootnote{The SDE-Net applied adversarial learning (AL) with noisy input samples during its training procedure. The training procedure of the AL is undertaken similarly to the test procedure of the open-set out-of-distribution detection task, as both of them add noises into the input samples. Thus, the AL can benefit the open-set out-of-distribution detection.} & $\ \ \boldsymbol{97.44\pm2.39}$\ {\tiny($\times$)}\ \  & \ \ $\boldsymbol{98.13\pm2.22}$\ {\tiny($\times$)}\ \  & \ \ $\boldsymbol{93.68\pm6.82}$\ {\tiny($\times$)}\ \  & \ \ $\boldsymbol{94.46\pm6.64}$\ {\tiny($\times$)}\ \  \\
    \ \ DS-UI (Ours) & \underline{$97.43\pm0.81$}\ {\tiny(\text{N/A})} & \underline{$97.50\pm0.76$}\ {\tiny(\text{N/A})} & \underline{$91.76\pm3.33$}\ {\tiny(\text{N/A})} & \underline{$91.72\pm3.40$}\ {\tiny(\text{N/A})} \\
    \hline
    \end{tabular}}
  \label{tab:oodis}
\end{table}

\begin{figure*}[!t]
    \centering
    \resizebox{1.01\linewidth}{!}{
    \begin{tabular}{l}
    \includegraphics[width=0.8\linewidth]{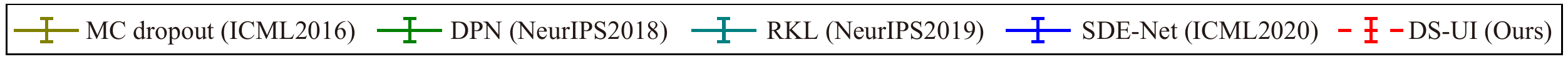} \\
    \begin{subfigure}[t]{0.249\linewidth}
        \centering
        \includegraphics[width=1\linewidth]{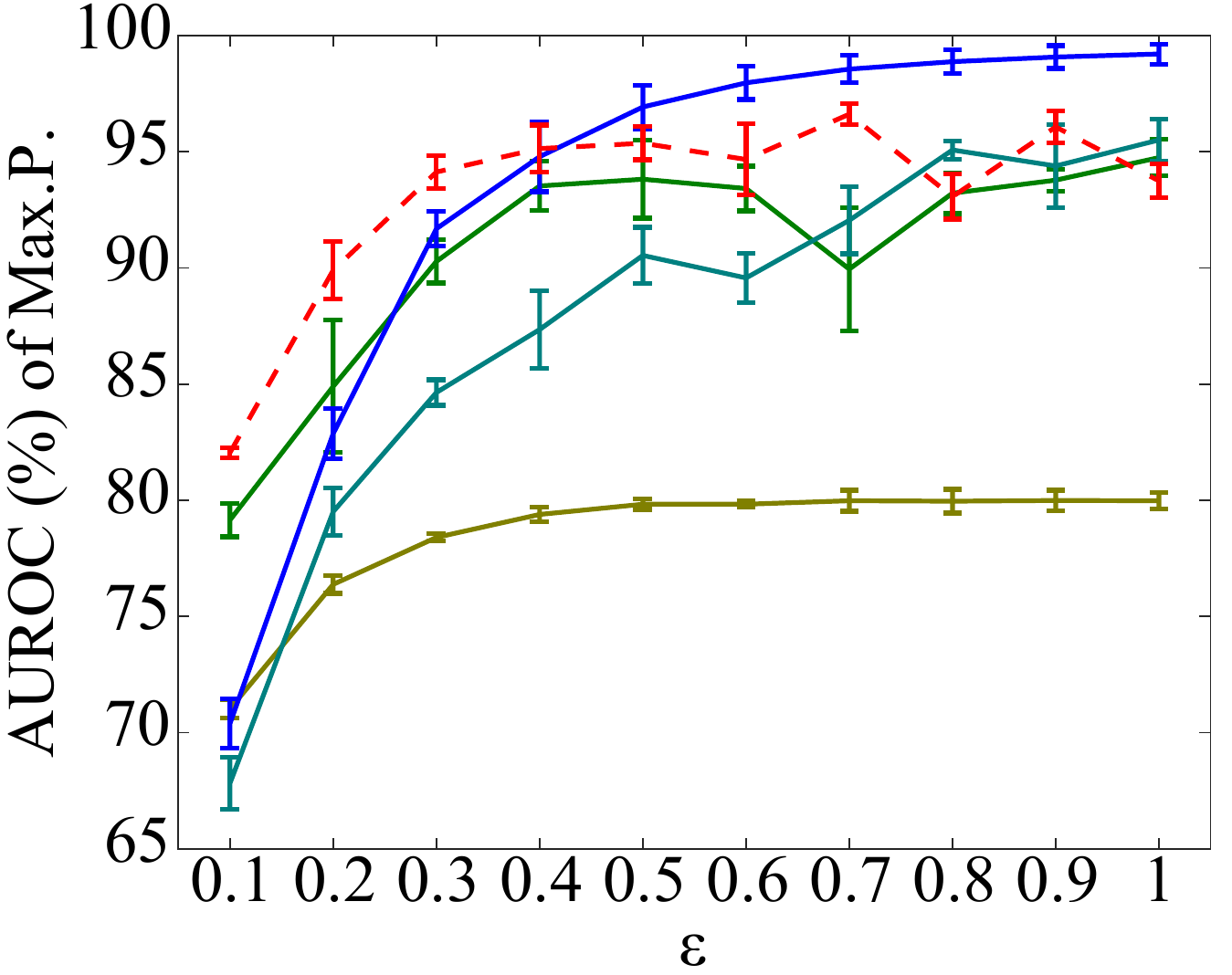}
        \subcaption{AUROC (\%) of Max.P.}
    \end{subfigure}
    \begin{subfigure}[t]{0.249\linewidth}
        \centering
        \includegraphics[width=1\linewidth]{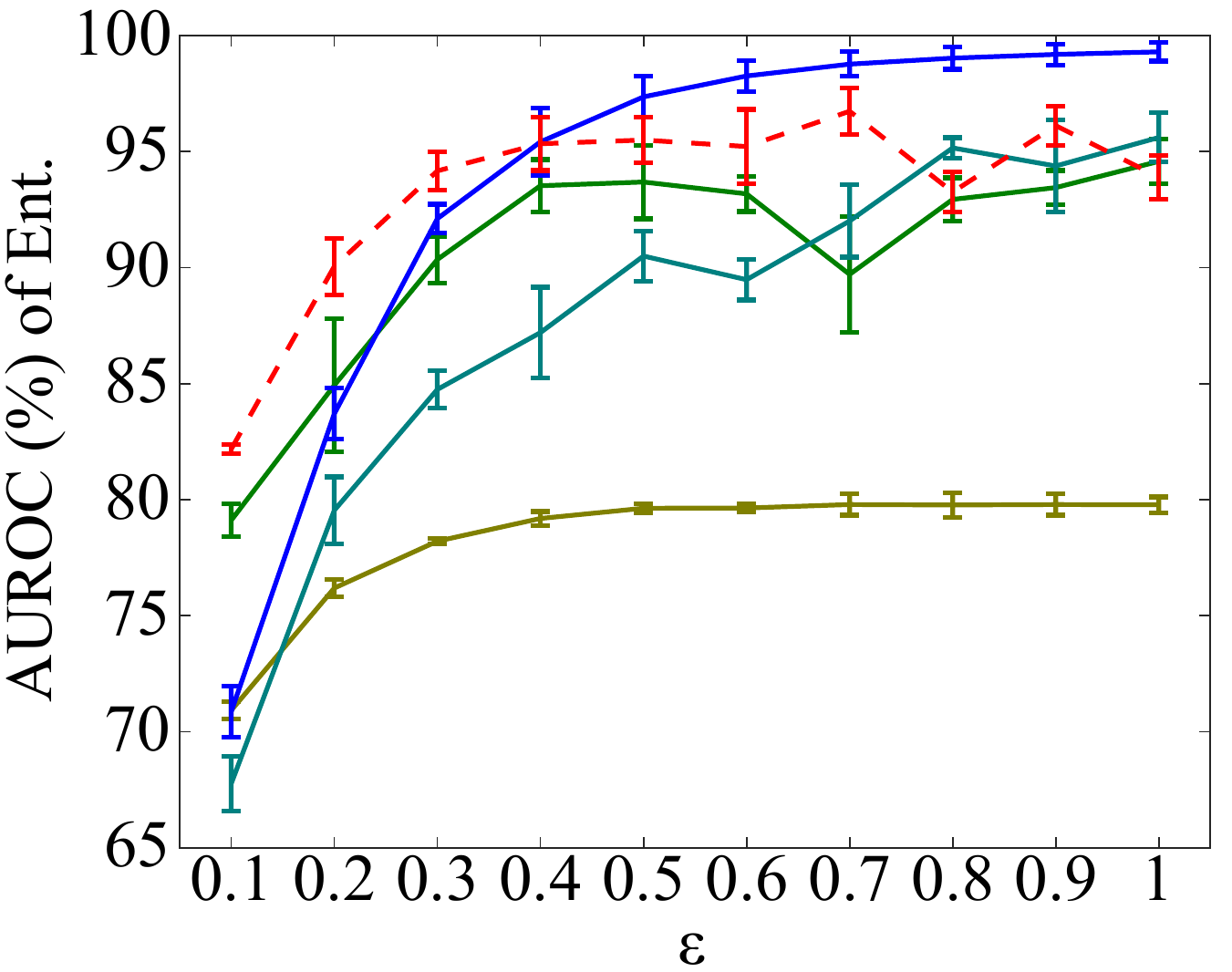}
        \subcaption{AUROC (\%) of Ent.}
    \end{subfigure}
    \begin{subfigure}[t]{0.249\linewidth}
        \centering
        \includegraphics[width=1\linewidth]{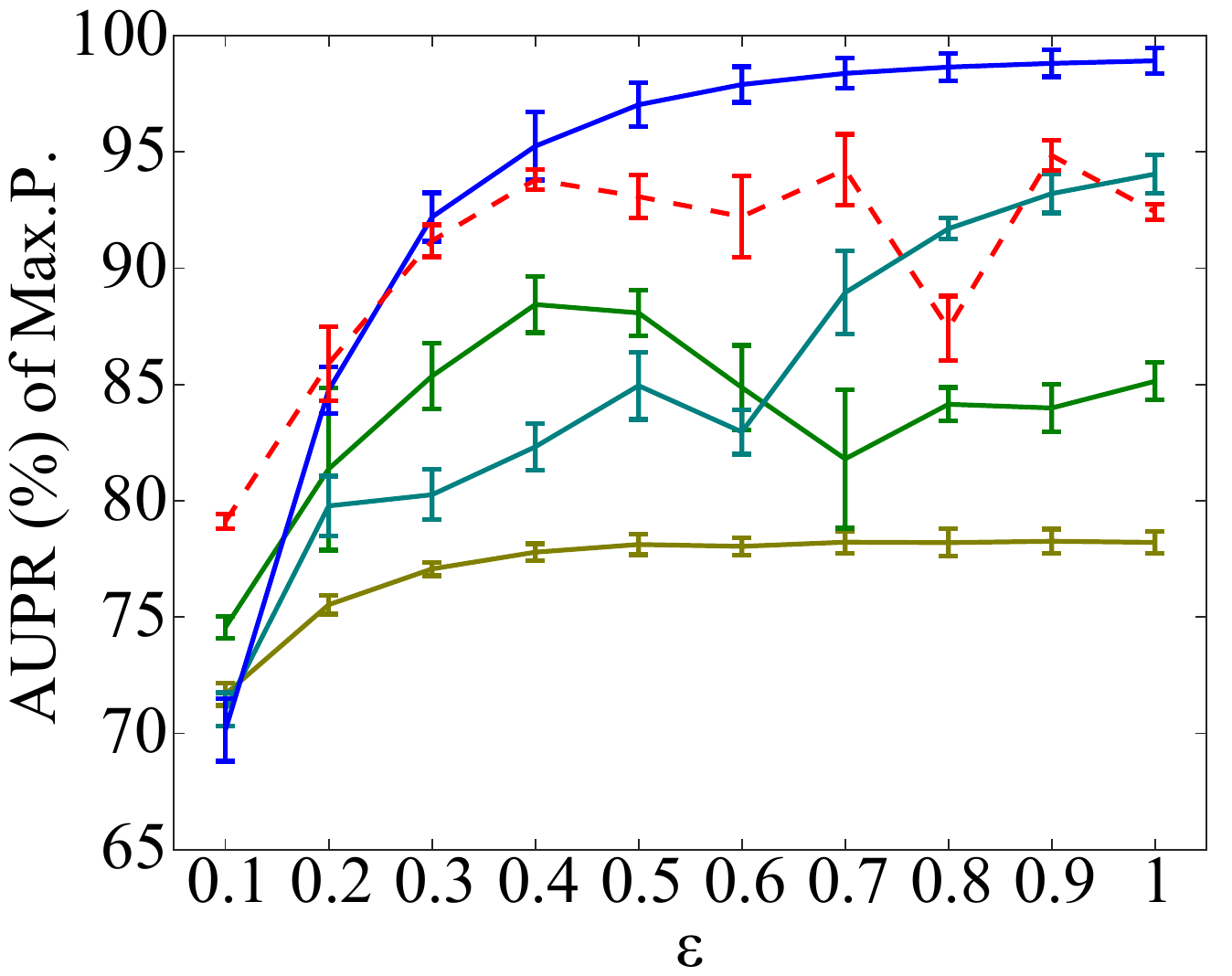}
        \subcaption{AUPR (\%) of Max.P.}
    \end{subfigure}
    \begin{subfigure}[t]{0.249\linewidth}
        \centering
        \includegraphics[width=1\linewidth]{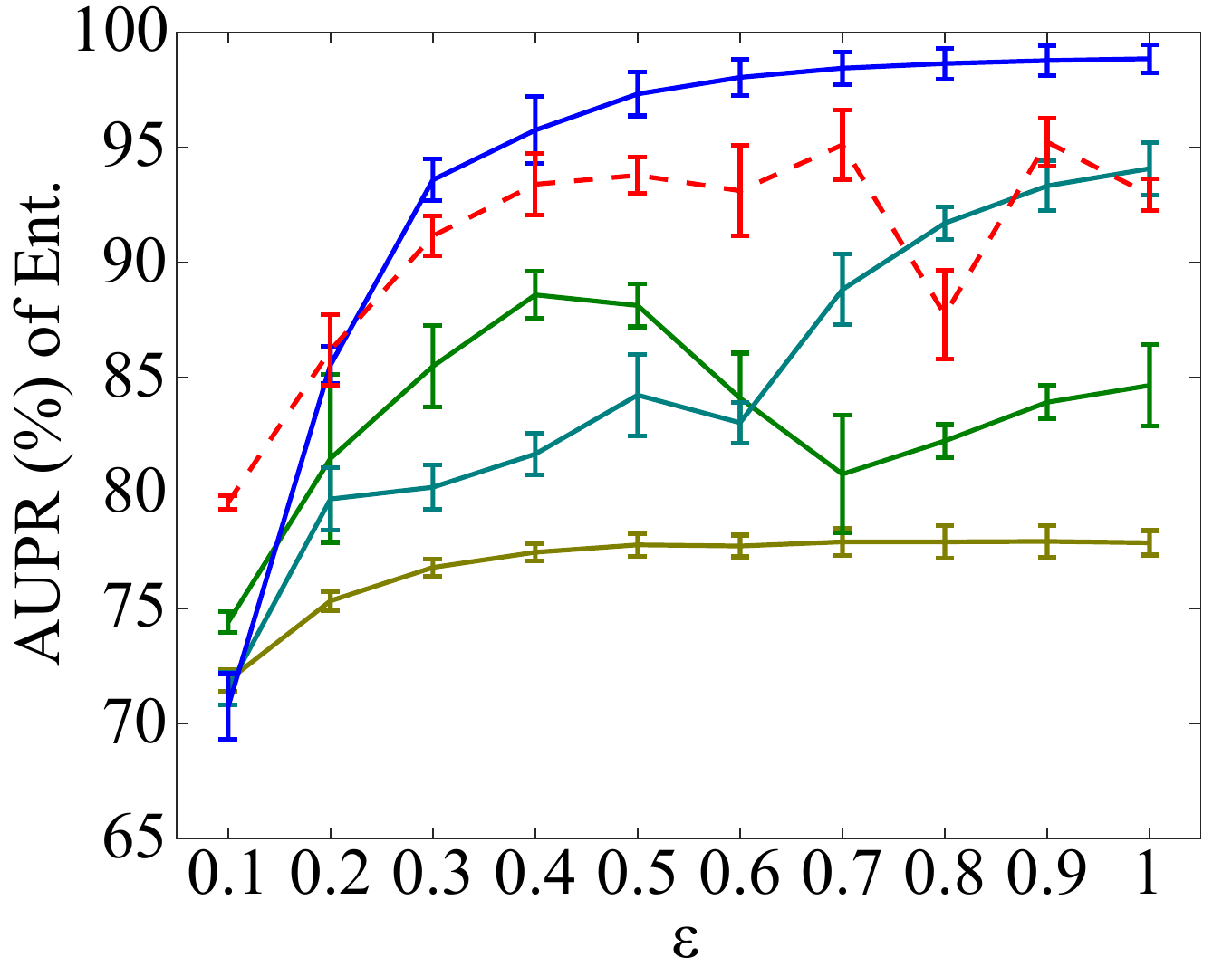}
        \subcaption{AUPR (\%) of Ent.}
    \end{subfigure}
    \end{tabular}}
    \caption{Performance of open-set out-of-distribution detection under FGSM attacks with ResNet$18$ on the CIFAR-$10$ dataset. $\varepsilon$ is the step size in the FGSM and selected in the set $\{\frac{n}{10}\}_{n=1}^{10}$. The error bars represent standard deviations of the values of the metrics. The dashed and solid lines in each subfigure present the DS-UI and the referred methods, respectively.}\label{fig:oodisadv}
\end{figure*}

\subsection{Misclassification Detection}\label{ssec:miscls}

The first important task in the UI is misclassification detection, which aims at detecting mispredicted samples in the test sets with uncertainty. Table~\ref{tab:cls} lists the image recognition accuracies with two backbones on four datasets. According to Table~\ref{tab:cls}, the DS-UI leads to the best performance in each case and achieves statistically significant performance improvement in most of the cases except the RKL with VGG$16$ on the CIFAR-$10$ dataset and ResNet$18$ on the CIFAR-$100$ dataset. Figure~\ref{fig:miscls} illustrates the experimental results in the misclassification detection. The DS-UI yields the best AUROC/AUPR in all the cases as well, and achieves statistically significant improvement in most of the cases. Therefore, we can conclude that the DS-UI is better for misclassification detection than the referred methods.

\subsection{Open-set Out-of-domain Detection}\label{ssec:ood}

We further evaluated the DS-UI in open-set out-of-domain detection. Different in-domain and out-of-domain dataset pairs were applied for the task, and the out-of-domain set in each pair was not used for training. Figure~\ref{fig:metricdistrib} shows the distributions of Max.P. and Ent. on the test sets of the ``CIFAR-$10\!\to$SVHN'' pair as an example. We can observe that distribution of out-of-domain samples is almost separated from those of in-domain classes, which means the DS-UI can effectively estimate uncertainty. Figure~\ref{fig:ood} shows that the DS-UI can surpass all the referred methods in most of the cases, except the AUPR of Ent.~on the ``TIM$\to$SVHN'' pair. Although the RKL outperforms the DS-UI in the case, there is no statistically significant difference between them, as the $p$-value of the unpaired Student's $t$-test is larger than $0.05$. In addition, the DS-UI obtains statistically significant improvement in most of the other cases, which shows the superiority of the DS-UI in the task.

In addition, we also evaluated the DS-UI following the settings in~\cite{perera2020generative}. The CIFAR-$10$ and the TIM datasets were divided into in-domain and out-of-domain sets, respectively. In Table~\ref{tab:ood2}, the best performance of the DS-UI under the metrics can be found on two datasets and the DS-UI achieves statistically significant improvement in all the cases on the TIM dataset and most of the cases on the CIFAR-$10$ dataset. The results show the remarkable ability of the DS-UI in the out-of-domain detection task.

\begin{figure*}[!t]
    \centering
    \resizebox{1.01\linewidth}{!}{
    \begin{tabular}{l}
    \includegraphics[width=0.75\linewidth]{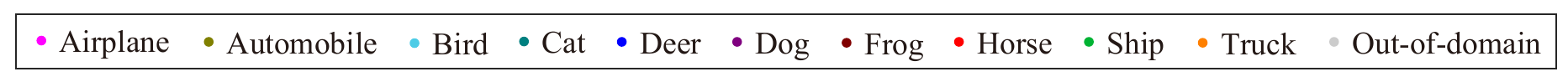} \\
    \begin{subfigure}[t]{0.249\linewidth}
        \centering
        \includegraphics[width=1\linewidth]{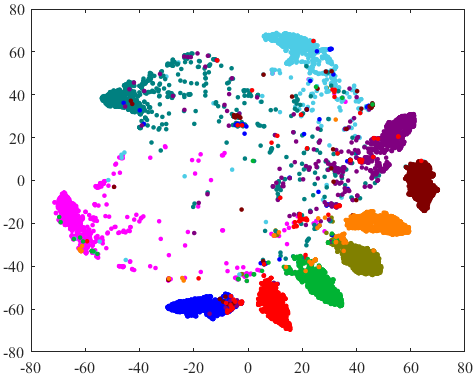}
        \subcaption{Baseline}
    \end{subfigure}
    \begin{subfigure}[t]{0.249\linewidth}
        \centering
        \includegraphics[width=1\linewidth]{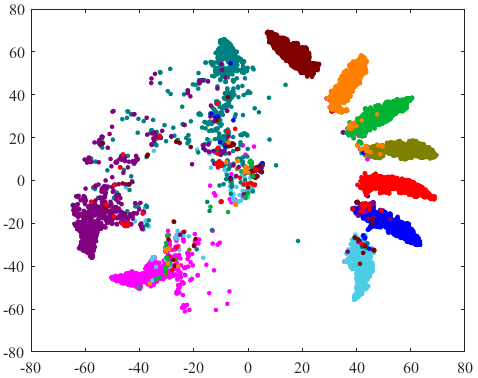}
        \subcaption{DS-UI}
    \end{subfigure}
    \begin{subfigure}[t]{0.249\linewidth}
        \centering
        \includegraphics[width=1\linewidth]{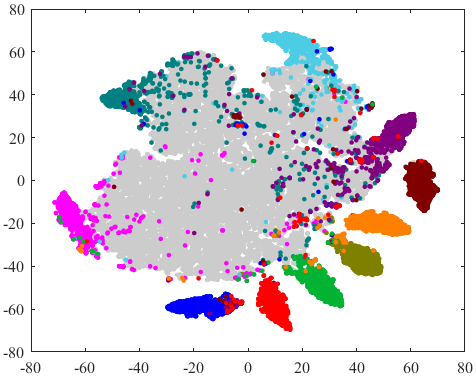}
        \subcaption{Baseline w/ OoD}
    \end{subfigure}
    \begin{subfigure}[t]{0.249\linewidth}
        \centering
        \includegraphics[width=1\linewidth]{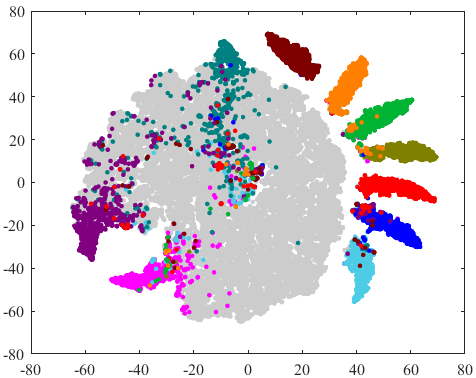}
        \subcaption{DS-UI w/ OoD}
    \end{subfigure}
    \end{tabular}}
    \caption{Visualizations of feature spaces of samples in the test sets of the baseline and the DS-UI with ResNet$18$ on the CIFAR-$10$ dataset as an example. The SVHN dataset is selected as the out-of-domain (OoD) dataset.}\label{fig:vis}
\end{figure*}

\subsection{Open-set Out-of-distribution Detection}\label{ssec:oodis}

We then evaluated the DS-UI in open-set out-of-distribution detection on a synthetic noise dataset. The dataset contains $10,000$ random images, where each pixel is independently sampled from a uniform distribution in $[0,1]$. Table~\ref{tab:oodis} shows the experimental results in open-set out-of-distribution detection between the CIFAR-$100$ dataset and the synthetic noise dataset. Although the DS-UI can only obtain the second best results under the metrics, no statistically significant difference is observed between the values of the evaluation metrics of the DS-UI and those of the SDE-Net (the $p$-values of the unpaired Student's $t$-test are all larger than $0.05$). The SDE-Net performs the best as it involves adversarial learning (AL) during its training procedure. This means it benefits from both the UI and the AL. In summary, the DS-UI works well and achieves comparable performance with the AL-based method (SDE-Net).

Furthermore, we evaluated the DS-UI under the adversarial attack, which can be considered as a distributional attack task, on the CIFAR-$10$ dataset. We introduced fast gradient-sign method (FGSM)~\cite{goodfellow2015explaining} as the attacker in the original input images on the test set. Treating the attacked images as the out-of-distribution samples, the adversarial attack task can be seen as an open-set out-of-distribution detection task. Parameter $\varepsilon$ in the FGSM presents the amplitude of the noises (or called the offset of distribution shift), which was selected in the set $\{\frac{n}{10}\}_{n=1}^{10}$~\cite{kong2020sde}. Figure~\ref{fig:oodisadv} shows the DS-UI obtains statistically significant improvement when $\varepsilon$ is small (adding minor noises) and even outperforms the AL-based SDE-Net. Although the SDE-Net can gain almost $99\%$ on all the four metrics when $\varepsilon$ is large (which are easier cases than the cases that minor noises are added), the DS-UI can perform comparably. Thus, the DS-UI can obtain superior ability in this task.

\subsection{Visualizations}\label{ssec:vis}

We conducted visualizations of the feature spaces of feature $\boldsymbol{z}$ of the baseline~\cite{hendrycks2017a} and the proposed DS-UI by t-distributed stochastic neighbor embedding (t-SNE)~\cite{maaten2008visualizing}, respectively, and show the results in Figure~\ref{fig:vis}. The ResNet$18$ model was used as the backbone, and the test sets of the CIFAR-$10$ and the SVHN datasets were used as the in-domain and the out-of-domain datasets, respectively. For the baseline in Figure~\ref{fig:vis}(a), all the classes are fused with each other and the inter-class margins are small, while the DS-UI in Figure~\ref{fig:vis}(b) obtains larger margins between most of the classes which is much better for the misclassification detection. Meanwhile, the intra-class distances of the DS-UI is also smaller than the baseline. More importantly, a clear and patent margin can be found between most of the in-domain classes and the out-of-domain samples in Figure~\ref{fig:vis}(d), even though some in-domain classes are partly confused with the out-of-domain samples. In the baseline model, the in-domain samples and the out-of-domain samples are more confusing with each other (Figure~\ref{fig:vis}(c)). It can be observed that the DS-UI can not only reduce intra-class distances, but also obtain much wider inter-class margins than the baseline model for both the misclassification detection and the open-set out-of-domain detection.

\section{Conclusions}\label{sec:conclusions}

In order to improve UI performance, DS-UI, a dual-supervised learning framework has been introduced to UI. Conventional UI methods commonly 
define uncertainty only on the outputs of DNNs. In the DS-UI, an MoGMM-FC layer that combines the classifier with an MoGMM was proposed to act as a probabilistic interpreter for the features of the DNNs. To enhance the learning ability of the MoGMM-FC layer, the DS-SGVB algorithm was proposed. It comprehensively considers both positive and negative samples to not only reduce the intra-class distances, but also enlarge the inter-class margins simultaneously. Experimental results show the proposed DS-UI outperforms the state-of-the-art UI methods in misclassification detection. In addition, we found the DS-UI can achieve statistically significant improvements in open-set out-of-domain/-distribution detection. Visualizations also support the superiority of the DS-UI for the learning ability enhancement.

In the future work, we plan to combine the DS-SGVB with other loss functions,~\emph{e.g.}, center loss~\cite{wen2016a} and A-softmax loss~\cite{liu2017sphereface}, and extend the MoGMM to the mixtures of mixture models with other distributions.

{\small
\bibliographystyle{ieee_fullname}
\bibliography{egbib}
}

\end{document}